\documentclass[10pt]{article}
\usepackage[english]{babel}

\usepackage{amsthm}
\usepackage{amsmath}
\usepackage[hidelinks]{hyperref} 
\usepackage{graphicx} 
\usepackage[utf8]{inputenc}
\usepackage{subcaption}
\usepackage{tikz}
\usetikzlibrary{bayesnet}
\usetikzlibrary{positioning}
\usetikzlibrary{decorations.text}
\usetikzlibrary{decorations.pathmorphing}
\usetikzlibrary{backgrounds}
\usepackage{bm}
\usepackage{multirow}
\usepackage{amssymb}
\usepackage{float}
\usepackage{units}

\usepackage[b5paper, twoside]{geometry}
\usepackage{titlesec}

\usepackage[longnamesfirst, round]{natbib}
\setcitestyle{authoryear,open={(},close={)}}

\numberwithin{equation}{section}
\theoremstyle{plain}

\theoremstyle{definition}

\definecolor{basecolor}{HTML}{C4C4C4}

\usepackage{authblk}

\newlength{\currentparskip}
\newlength{\currentparindent}
\newenvironment{myabstract}[1]
    {
    \setlength{\currentparskip}{\parskip}
    \setlength{\currentparindent}{\parindent}
    \begin{center}
    \begin{minipage}[]{#1}
    \setlength{\parskip}{\currentparskip}
    \setlength{\parindent}{\currentparindent}
    \small
    \begin{center} \textbf{Abstract} \end{center} \vspace*{-0.5em} \vspace*{-\currentparskip}
    }
    {
    \end{minipage}
    \end{center}
    }
    
\newenvironment{mykeywords}[1]
    {
    \vspace*{-1.5em}
    \begin{center}
    \begin{minipage}[]{#1}
    \small
    \noindent \textbf{Keywords:}
    }
    {
    \end{minipage}
    \end{center}
    }

\begin{document}

\graphicspath{{./figures/}}

%%%%%%%%%%%%%%%%%%%%%%%%%%

\title{Bayesian Calibration of Imperfect Computer Models
using Physics-Informed Priors}

\author[1]{Michail Spitieris}
\author[1]{Ingelin Steinsland}

\affil[1]{Department of Mathematical Sciences, NTNU, Norway}
\date{}

\maketitle 

%%%%%%%%%%%%%%%%%%%%%%%%%%%%%

\begin{myabstract}{0.85\linewidth}

\noindent
We introduce a computational efficient data-driven framework suitable for quantifying the uncertainty in physical parameters and model formulation of computer models, represented by differential equations. We construct physics-informed priors, which are multi-output GP priors that encode the model's structure in the covariance function. This is extended into a fully Bayesian framework that quantifies the uncertainty of physical parameters and model predictions. Since physical models often are imperfect descriptions of the real process, we allow the model to deviate from the observed data by considering a discrepancy function. 
For inference, Hamiltonian Monte Carlo is used.
Further, approximations for big data are developed that reduce the computational complexity from $\mathcal{O}(N^3)$ to $\mathcal{O}(N\cdot m^2),$ where $m \ll N.$
Our approach is demonstrated in simulation and real data case studies where the physics are described by time-dependent ODEs describe  (cardiovascular models) and space-time dependent PDEs (heat equation). In the studies, it is shown that our modelling framework can recover the true parameters of the physical models in cases where 1) the reality is more complex than our modelling choice and 2) the data acquisition process is biased while also producing accurate predictions. Furthermore, it is demonstrated that our approach is computationally faster than traditional Bayesian calibration methods.

\end{myabstract}

\begin{mykeywords}{0.85\linewidth}
Gaussian process, 
model discrepancy, 
physics-informed prior, 
inverse problem, 
HMC, 
arterial Windkessel, 
Heat equation, 
physics-informed ML
\end{mykeywords}

\section{Introduction}

Physical models are  mathematical representations of the phenomenon under study and are commonly described by (systems of) differential equations. They are usually deduced from first principles and aim to describe the underlying physics explicitly. 
In contrast to purely data-driven models, they allow predictions in regions where we do not have observed data (extrapolation). For example, we can predict future evolution of heat in a material at time $t_\text{pred}$ given observed data up to time $t_{\text{obs}},$ where $t_{\text{obs}}<t_\text{pred}.$ To enable model predictions, we have to estimate a set of unknown parameters based on observations. Conventional methods for estimating the unknown parameters use the observed data in curve fitting algorithms. However, even under the best set of these parameters, the fit to the observed data often suffers from systematic discrepancies.

We consider the situation that we have a possible imperfect model based on linear differential equations, and that the aim is to estimate these parameters based on noisy data. Such parameters often have a concrete scientific interpretation. For example, in our case studies using the Windkessel model, the hemodynamical parameters, arterial compliance, and total peripheral resistance can provide insights into the development of hypertension.
Hence, the parameters are of interest on their own. Another common situation is that predictions are the main interest, but to be able to use the physical models for predictions, the unknown parameters are needed. 

In this paper, we propose and demonstrate a new method for computationally efficiently estimating the model parameters, including possible prior knowledge and the possibility for model discrepancy. We achieve this by combining the framework of Bayesian calibration for accounting for imperfect models \citep{kennedy2001bayesian} with physics-informed priors for linear differential equations \citep{raissi2017machine}.

\vspace{0.2cm}

\noindent \textbf{Notation} We consider physical models formulated as linear parametric differential equations $\mathcal{L}_{x}^{\boldsymbol{\phi}}u(x) = f(x),$ where $\mathcal{L}$ is the linear differential operator and $\boldsymbol{\phi}=(\phi_1,\ldots,\phi_p)$ is the vector of physical parameters. 
For example, for the first order non-homogeneous differential equation, $\phi_1\frac{du(x)}{dx} + \phi_2^{-1}u(x)=f(x)$ we want to estimate the parameters $\phi_1$ and $\phi_2$ (an example of first order differential equation can be found in Section \ref{sec:WK_synth}). We denote the observed data of the function $u$ at $\mathbf{X}_u = (X_{u_1}, \ldots, X_{u_{n_u}})$ as $\mathbf{y}_u = (y_{u_1},\ldots, y_{u_{n_u}})$ and similarly for the function $f$ at $\mathbf{X}_f = (X_{f_1}, \ldots, X_{f_{n_f}})$ as $\mathbf{y}_f = (y_{f_1},\ldots, y_{f_{n_f}}),$ where $n_u$ and $n_f$ is the number of observed data for the functions $u$ and $f,$ respectively.

\subsection{Accounting for model discrepancy using Bayesian calibration} \label{sec:KOH_intro}
It has been twenty years since the seminal paper of \citet*{kennedy2001bayesian} (KOH) where they introduced the idea of Bayesian calibration by accounting for model discrepancy. In their model formulation, they added a functional discrepancy term, $\delta(x)$ to account for the model-form uncertainty which arises from a low-fidelity  physical model. More specifically, they modelled the noise corrupted observed data, $y,$ by the physical model, $\eta$ and the systematic model discrepancy as $y(x)=\eta(x,\boldsymbol \phi)+\delta(x)+\varepsilon,$ where $x$ is the observed inputs and $\boldsymbol \phi$ is the set of (unknown) physical parameters. A flexible Gaussian process (GP) prior \citep{williams2006gaussian} was used for the model discrepancy, $\delta(x)\sim GP(0,K_{\delta}(x,x')),$ where $K$ denotes the covariance function. 

% \textbf{Applications of BC}
The KOH formulation has been applied in many fields of science, including engineering \citep{bayarri2009predicting}, hydrology \citep{reichert2009analyzing}, ecology \citep{arhonditsis2008bayesian}, health sciences \citep{strong2012managing, spitieris2022bayesian}, biology \citep{henderson2009bayesian}, climate modelling \citep{forest2008inferring, goldstein2009reified, salter2019uncertainty} and astrophysics \citep{habib2007cosmic}.

Often the main challenge of this approach is that the numerical simulator of the physical model $\eta$ is computationally expensive, and KOH replaced the model with an emulator \citep{sacks1989design} which is a statistical approximation to the model. A typical choice for an emulator is a GP model trained on the numerical simulator runs created according to an experimental design on $[X,\boldsymbol \phi]-$space. For inference, simulation data of size $N$ and observed data of size $n$ are used, where $N\gg n.$ However, the physical model outputs are usually functional, and more than one, hence the $\mathcal{O}((N+n)^3)$ computational complexity of the GP model can be prohibitive in such cases.

\citet{higdon2004combining} utilized this formulation for models that numerical simulators are not expensive to evaluate, and therefore there is no need for constructing an emulator. To deal with the computational complexity of the emulator in the case of multivariate and time-dependent outputs, principal components analysis (PCA) has been used to reduce the dimensionality of the problem \citep{higdon2008computer,higdon2008bayesian}. Other approaches involve the modification of the GP emulator, for example, through a composite likelihood \citep{chang2015composite}, local approximate GP regression \citep{gramacy2015local} and basis representations \citep{bayarri2007computer, chang2019computer}. Recent advances in Deep GPs \citep{damianou2013deep} with random feature expansion \citep{cutajar2017random} have allowed more complex modelling structures \citep{marmin2022deep}.

Furthermore, \citet*{brynjarsdottir2014learning} showed through a motivating example that not accounting for model discrepancy in a low-fidelity physical model can lead to biased and over-confident physical parameter estimates. 

\subsection{Physics-informed priors}
Let  $u(x)\sim GP(0, K_{uu}(x,x')),$ denote a GP with mean 0 and kernel $K_{uu}(x,x')$ where $K_{uu}(x,x')$ is the covariance between the process $u(x)$ at location $x$ and $x'$, $K_{uu}(x,x') = Cov(u(x), u(x'))$, which typically involves parameters that we will denote $\theta$, but are suppressed for now.
A key property that enables the construction of physics-informed priors, is that the derivatives of a Gaussian process are also a Gaussian process \citep[Theorem 2.2.2]{adler2010geometry}. 
We then have that
\begin{equation} 
   \textrm{Cov}\left(u(x), \frac{\partial u(x')}{\partial x'}\right) = \frac{\partial K_{uu}(x,x')}{\partial x'}
   \quad \text{and} \quad 
   \textrm{Cov}\left(\frac{\partial u(x)}{\partial x}, \frac{\partial u(x')}{\partial x'}\right) = \frac{\partial^2 K_{uu}(x,x')}{\partial x \partial x'} \label{eq:adler}.
\end{equation}
Eq. \ref{eq:adler} is valid only if the covariance function is differentiable, thus a convenient choice can be the squared exponential kernel, $K(x,x') = \sigma^2{ \exp\left(- 0.5\,{ \left( {\frac {x-x'}{l}} \right) }^{2}\right)}$ where $l$ is a parameter scaling the strength of the dependency.
To build physics-informed priors for linear differential equations, we follow the idea of \cite{raissi2017machine} where they assume that $u(t)\sim GP(0, K_{uu}(x,x')),$ and then by using eq. \ref{eq:adler} we have that $f(t) \sim GP(0, K_{ff}(x,x')),$ where $K_{ff}(x,x')=\mathcal{L}_{x}^{\boldsymbol\phi}\mathcal{L}_{x'}^{\boldsymbol\phi}K_{uu}(x,x')$, and also the covariances between $u$ and $f$ are 
$K_{uf}(x,x') =\mathcal{L}_{x'}^{\boldsymbol\phi}K_{uu}(x,x')$ and 
$K_{fu}(x,x') =\mathcal{L}_{x}^{\boldsymbol\phi}K_{uu}(x,x').$ Note that the covariance functions $K_{ff}$ and $K_{uf}$ are functions of the physical parameters $\boldsymbol \phi$. The advantage of this approach is that we have built a multi-output GP (of $u$ and $f$) which bypass the need to solve the differential equation numerically, which can be computationally inefficient and also the physical parameters $\boldsymbol{\phi}$ are now hyperparameters of the kernel.  \cite{raissi2017machine} obtained point estimates of the physical parameters by maximizing the marginal log-likelihood. 

\vspace{0.7cm}

This work is motivated by a medical Digital Twin for prevention and treatment of hypertension (or high blood pressure). 
Physical models of the cardiovascular system allow estimating physical parameters that are important to Hypertension and can not be measured directly. 
For example, a low-fidelity model of the cardiovascular system, the Windkessel model \citep{westerhof2009arterial} (introduced in Section \ref{sec:WK_synth}), is a linear differential equation linking blood pressure and blood flow. It has two physical parameters, arterial compliance $C$ and resistance $R.$ 
These are unknown in practice but can be estimated by fitting the physical model to the observed blood pressure and inflow data, measured by sensors. 

We know that low-fidelity cardiovascular models are imperfect mathematical representations of the real process, and if we do not account for model discrepancy, the parameter estimates are biased \citep{brynjarsdottir2014learning}. 
Therefore, we want to incorporate a model discrepancy term in the model formulation to account for the model's missing physics, as suggested by \citet{kennedy2001bayesian}. 
The differential equations typically use numerical solvers to simulate data from the model, and their computational cost can be prohibitive for an MCMC scheme. For this reason, KOH built an emulator, which is a GP model trained on simulator data. More specifically, to obtain data for the emulator, we run the simulator on an experimental design on the input and physical parameter space (for example, a Latin hypercube design). The challenges for employing this approach in a Digital Twin technology are
1) in order to run the simulator, initial and boundary conditions might be needed, which might not be known in practice,
2) the KOH approach and methods discussed in Section \ref{sec:KOH_intro} utilize two sources of information, the $N$ simulator data and $n$ observed data and have complexity $\mathcal{O}((N+n)^3),$ which might be prohibitive for Digital Twin technologies,
3) finding an appropriate experimental design can be a challenging task that is hard to automate, and also the design can significantly affect the result of the KOH approach.

\vspace{0.7cm}

\noindent \textbf{Contributions} (i) Contributions from physics-informed priors point of view: We extend the idea of physics-informed priors in a fully Bayesian framework that allows for quantifying the uncertainty in physical parameters. Further, we  incorporate a functional model discrepancy in the physics-informed prior formulation to account for imperfect models. \\
(ii) Contribution from Bayesian calibration point of view: We replace the computationally expensive emulator by the physics-informed prior and this reduces the complexity from $\mathcal{O}((N+n)^3)$ to $\mathcal{O}(n^3),$ since the model is evaluated only on the observed data. \\
(iii) Modelling flexibility contribution: Using the physics-informed prior in a fully Bayesian framework allows for more flexible modelling and we demonstrate this flexibility by considering a case where the data acquisition process is biased. \\
(iv) Approximations for big data: We derive approximations for our models that reduce the computational cost from $\mathcal{O}(N^3)$ to $\mathcal{O}(N\cdot m^2),$ where $m \ll N.$ 
\vspace{0.5cm}

The remainder of the paper is organized as follows. In Section \ref{sec:method}, we formally define the physics-informed prior models. First, the fully Bayesian extension (Section \ref{sec:FullBayes}), then the Bayesian calibration framework with model discrepancy (Section \ref{sec:BCPI_main}) and finally the model for biased data acquisition process (Section \ref{sec:PI_biased_sensors}). In Section \ref{sec:WK_synth}, we consider a simulation study with the Windkessel models, which are time-dependent differential equations where we account for model discrepancy. In Section \ref{sec:HF}, we use the Heat equation, which is space and time-dependent differential equation, where we consider a simulation study with biased sensor data. In Section \ref{sec:real_data}, we demonstrate a real data case study with the Windkessel physics-informed prior model. In Section \ref{sec:comparison}, we compare the proposed approach with methods it improves.
In Section \ref{sec:bigdata}, we derive approximations for our method, which reduce the computational complexity to $\mathcal{O}(N\cdot m^2).$
Finally, in Section \ref{sec:conc}, we discuss the results and further work. The code to replicate all the results in the paper is available at
\url{https://github.com/MiSpitieris/BC-with-PI-priors}.

% -----------------------------------------------------------------------
% -----------------------------------------------------------------------
% -----------------------------------------------------------------------
\section{Bayesian calibration with physics-informed priors}\label{sec:method}
In this section we introduce the Bayesian calibration framework for computer models described by linear parametric differential equations, $\mathcal{L}_{x}^{\boldsymbol\phi}u(x) = f(x),$ using physics-informed priors. In Section \ref{sec:FullBayes}, we extend this in a fully Bayesian framework. To account for imperfect physical models the model formulation is extended  incorporating a functional model discrepancy in Section \ref{sec:BCPI_main}. Section \ref{sec:PI_biased_sensors} introduces the model formulation for biased data.

%--------------------------------------------------------------------------------------
% 2.1
\subsection{Fully Bayesian analysis with physics-informed priors}\label{sec:FullBayes}
For the linear differential equation, $\mathcal{L}_{x}^{\boldsymbol\phi}u(x) = f(x)$ we follow \cite{raissi2017machine} and build physics-informed priors assuming  $u(x)\sim GP(\mu_u(x), K_{uu}(x,x')).$ Unlike \cite{raissi2017machine} we also include a mean function, $\mu_u(x\mid \boldsymbol{\beta}),$ and this results in $\mu_f(x\mid \boldsymbol{\beta, \phi}) = \mathcal{L}_{x}^{\boldsymbol\phi}\mu_u(x)$, where $\boldsymbol{\beta}$ is the vector of the mean function parameters. The observed data, $y_u$ and $y_f$ are modelled by the physics-informed prior with i.i.d. Gaussian noise, $\varepsilon_u\sim N(0, \sigma_u^2)$ and $\varepsilon_f\sim N(0, \sigma^2_f),$ respectively, 

\begin{equation*} 
  \begin{split}
    y_u & = u(x_u) + \varepsilon_u,\\
	y_f &= f(x_f) + \varepsilon_f.
  \end{split}
\end{equation*}
This results in the following multi-output GP 

\begin{equation}
  p(\mathbf{y}\mid \boldsymbol \theta,  \boldsymbol \phi, \sigma_u, \sigma_f) = \mathcal{N}(\boldsymbol \mu, \mathbf{K} + \mathbf{S}) \label{noDisc}
\end{equation}
where 
$
\bf{y} = \begin{bmatrix}  \bf{y}_u \\  \bf{y}_f \end{bmatrix}
$,
$
\mathbf{K} =
\begin{bmatrix}
K_{uu}(\mathbf{X}_u, \mathbf{X}_u \mid \boldsymbol \theta)  & K_{uf}(\mathbf{X}_u, \mathbf{X}_f \mid \boldsymbol \theta,  \boldsymbol \phi)\\
K_{fu}(\mathbf{X}_f, \mathbf{X}_u \mid \boldsymbol \theta,  \boldsymbol \phi) & K_{ff}(\mathbf{X}_f, \mathbf{X}_f \mid \boldsymbol \theta,  \boldsymbol \phi)
\end{bmatrix}
$,\\
$
\mathbf{S} =
\begin{bmatrix}
\sigma_u^2 I_u & 0\\
0 & \sigma_f^2 I_f
\end{bmatrix} 
$, 
$
\boldsymbol \mu =
\begin{bmatrix}
\boldsymbol \mu_u(\mathbf{X}_u \mid \boldsymbol \beta)\\
\boldsymbol \mu_f(\mathbf{X}_f \mid \boldsymbol{\beta,\phi})
\end{bmatrix}
$  
and $\theta$ is the parameters of the kernel of the GP prior for $\boldsymbol u.$

We assign priors to the physical model parameters $\boldsymbol \phi$ that reflect underlying scientific knowledge and also assign priors to the mean, kernel and noise parameters. For convenience,  we denote all the parameters collectively $\boldsymbol \xi = (\boldsymbol \phi, \boldsymbol \beta, \boldsymbol \theta, \sigma_u, \sigma_f).$ To sample the posterior distribution of $ \boldsymbol \xi$ standard sampling methods can be used. In this paper, we use Hamiltonian Monte Carlo (HMC) sampling and more specifically, the No U-Turn Sampler (NUTS) \citep{hoffman2014no} variation implemented in the probabilistic programming language STAN \citep{carpenter2017stan}.

Suppose now that we want to make predictions at new points $X^*_u,$ $u(X^*_u) = \mathbf{u}^*.$ The conditional distribution $p(\mathbf{u}^* \mid \mathbf{X}^*_u, \mathbf{X}, \mathbf{y}, \boldsymbol{\xi})$ is multivariate Gaussian (see Appendix \ref{app:PE} for derivation) and more specifically

\begin{align*}
p(\mathbf{u}_* \mid \mathbf{X}_u^*, \mathbf{X}, \mathbf{y}, \boldsymbol{\xi}) &= \mathcal{N}(\boldsymbol{\mu}_u^*, \boldsymbol{\Sigma}_u^*) \\
\boldsymbol{\mu_u^*} &= \mu_u\mathbf{(X_u^*)}+ \mathbf{V}_u^*{^T} (\mathbf{K}+\mathbf{S})^{-1} (\mathbf{y}-\boldsymbol{\mu})\\
\boldsymbol{\Sigma_u^*} &= K_{uu}(\mathbf{X}_u^*,\mathbf{X}_u^*) - \mathbf{V}_u^*{^T} (\mathbf{K}+\mathbf{S})^{-1} \mathbf{V}_u^*,
\end{align*}
where $\mathbf{V}_u^*{^T} = \begin{bmatrix}  K_{uu}(\mathbf{X}_u^*,\mathbf{X}_u) &  K_{uf}(\mathbf{X}_u^*,\mathbf{X}_f) \end{bmatrix}.$\\
Similarly, at new points  $\mathbf{X}_f^*,$ for the predictions $f(\mathbf{X}_f^*)=\mathbf{f^*}$ we have that

\begin{align*}
p(\mathbf{f}_* \mid \mathbf{X}_f^*, \mathbf{X}, \mathbf{y}, \boldsymbol{\xi}) &= \mathcal{N}(\boldsymbol{\mu}_f^*, \boldsymbol{\Sigma}_f^*) \\
\boldsymbol{\mu_f^*} &= \mu_f\mathbf{(X_f^*)}+ \mathbf{V}_f^*{^T} (\mathbf{K}+\mathbf{S})^{-1} (\mathbf{y}-\boldsymbol{\mu})\\
\boldsymbol{\Sigma_f^*} &= K_{ff}(\mathbf{X}_f^*,\mathbf{X}_f^*) - \mathbf{V}_f^*{^T} (\mathbf{K}+\mathbf{S})^{-1} \mathbf{V}_f^*,
\end{align*}
where $\mathbf{V}_f^*{^T} = \begin{bmatrix}  K_{fu}(\mathbf{X}_f^*,\mathbf{X}_u) &  K_{ff}(\mathbf{X}_f^*,\mathbf{X}_f) \end{bmatrix}.$

%-----------------------------------------------------------------------------
% 2.2
\subsection{Physics-informed priors for imperfect models} \label{sec:BCPI_main}
Physical models are often imperfect representations of reality. To incorporate this in the model formulation, we follow \cite{kennedy2001bayesian} and include a functional model discrepancy. For simplicity, we assume discrepancy only on the function $u(x)$ and we get the following model formulation

\begin{equation*} 
  \begin{split}
    y_u & = u(x_u) + \delta_u(x_u) + \varepsilon_u, \text{ where } \delta_u(x) \sim GP(0, K_{\delta_u}(x,x'))\\
	y_f &= f(x_f) + \varepsilon_f.
  \end{split}
\end{equation*}
We follow Section \ref{sec:FullBayes} and assume Gaussian i.i.d. noise and physics-informed priors.  This results in the following multi-output GP 

\begin{equation}
	p(\mathbf{y}\mid \boldsymbol \theta, \boldsymbol\theta_{\delta_u}, \boldsymbol \phi, \sigma_u, \sigma_f) = \mathcal{N}(\mathbf{\mu}, \mathbf{K}_{\text{disc}}+\mathbf{S}) \label{Disc}
\end{equation}
where
$
\bf{y} = \begin{bmatrix}  \bf{y}_u \\  \bf{y}_f \end{bmatrix}
$,
$
\mathbf{K}_\text{disc} =
\begin{bmatrix}
K_{uu}(\mathbf{X}_{u}, \mathbf{X}_{u} \mid \boldsymbol \theta) + K_\delta (\mathbf{X}_u,\mathbf{X}_u\mid \boldsymbol{\theta}_\delta)  & K_{uf}(\mathbf{X}_u, \mathbf{X}_f \mid \boldsymbol \theta,  \boldsymbol \phi)\\
K_{fu}(\mathbf{X}_f, \mathbf{X}_u \mid \boldsymbol \theta,  \boldsymbol \phi) & K_{ff}(\mathbf{X}_{f}, \mathbf{X}_{f} \mid \boldsymbol \theta,  \boldsymbol \phi)
\end{bmatrix}
$,\\
$
\mathbf{S} =
\begin{bmatrix}
\sigma_u^2 I_u & 0\\
0 & \sigma_f^2 I_f
\end{bmatrix} 
$ and 
$
\boldsymbol \mu =
\begin{bmatrix}
\boldsymbol \mu_u(\mathbf{X}_u \mid \boldsymbol \beta)\\
\boldsymbol \mu_f(\mathbf{X}_f \mid \boldsymbol{\beta,\phi})
\end{bmatrix}.
$ 

Considering the covariance matrix $K_{\text{disc}}$  the only change compared to the covariance matrix of \eqref{noDisc} is an added term corresponding to the covariance matrix for the discrepancy for the $y_u$ part. We have augmented the parameters in the vector $\boldsymbol{\theta}_\delta.$ 
As in Section \ref{sec:FullBayes}, we use a fully Bayesian approach where we assign prior distributions to all unknown parameters denoted jointly as $\boldsymbol{\xi}_{\text{disc}} = (\boldsymbol \phi, \boldsymbol \beta, \boldsymbol \theta, \boldsymbol{\theta}_\delta,  \sigma_u, \sigma_f)$ and inference is performed using HMC.

In order to make predictions at new points $X^*_u,$ $u(X^*_u) = \mathbf{u}^*$ we use that the conditional distribution $p(\mathbf{u}^* \mid \mathbf{X}^*_u, \mathbf{X}, \mathbf{y}, \boldsymbol{\xi}_{\text{disc}})$ is multivariate Gaussian and more specifically

\begin{align*}
p(\mathbf{u}_* \mid \mathbf{X}_u^*, \mathbf{X}, \mathbf{y}, \boldsymbol{\xi}_\delta) &= \mathcal{N}(\boldsymbol{\mu}_u^*, \boldsymbol{\Sigma}_u^*) \\
\boldsymbol{\mu_u^*} &= \mu_u\mathbf{(X_u^*)}+ \mathbf{V}_u^*{^T} (\mathbf{K_{\text{disc}}}+\mathbf{S})^{-1} (\mathbf{y}-\boldsymbol{\mu})\\
\boldsymbol{\Sigma_u^*} &= K_{uu}(\mathbf{X}_u^*,\mathbf{X}_u^*)+K_{\delta}(\mathbf{X}_u^*,\mathbf{X}_u^*) - \mathbf{V}_u^*{^T} (\mathbf{K_{\text{disc}}}+\mathbf{S})^{-1} \mathbf{V}_u^*,
\end{align*}
where $\mathbf{V}_u^*{^T} = \begin{bmatrix}  K_{uu}(\mathbf{X}_u^*,\mathbf{X}_u)+K_{\delta}(\mathbf{X}_u^*,\mathbf{X}_u) &  K_{uf}(\mathbf{X}_u^*,\mathbf{X}_f) \end{bmatrix}.$ In comparison to model \ref{noDisc}, the predictive equations now includes the  discrepancy $\delta(x)$ which models the missing physics.

The conditional distribution $p(\mathbf{f}_* \mid \mathbf{X}_f^*, \mathbf{X}, \mathbf{y}, \boldsymbol{\xi})$ is multivariate Gaussian and more specifically

\begin{align*}
p(\mathbf{f}_* \mid \mathbf{X}_f^*, \mathbf{X}, \mathbf{y}, \boldsymbol{\xi}_{\delta}) &= \mathcal{N}(\boldsymbol{\mu}_f^*, \boldsymbol{\Sigma}_f^*) \\
\boldsymbol{\mu_f^*} &= \mu_f\mathbf{(X_f^*)}+ \mathbf{V}_f^*{^T} (\mathbf{K_{\text{disc}}}+\mathbf{S})^{-1} (\mathbf{y}-\boldsymbol{\mu})\\
\boldsymbol{\Sigma_f^*} &= K_{ff}(\mathbf{X}_f^*,\mathbf{X}_f^*) - \mathbf{V}_f^*{^T} (\mathbf{K_{\text{disc}}}+\mathbf{S})^{-1} \mathbf{V}_f^*,
\end{align*}
where $\mathbf{V}_f^*{^T} = \begin{bmatrix}  K_{fu}(\mathbf{X}_f^*,\mathbf{X}_u) &  K_{ff}(\mathbf{X}_f^*,\mathbf{X}_f) \end{bmatrix}.$
The prediction equations for $\mathbf{f}^*$ are similar to those presented in Section \ref{sec:Biased_model}.
%but know the covariance matrix $\mathbf{K}$ has been replaced by $\mathbf{K}_\text{disc}.$ 
For more details on the derivation of equations $p(\mathbf{u}_* \mid \mathbf{X}_u^*, \mathbf{X}, \mathbf{y}, \boldsymbol{\xi}_\delta)$ and $p(\mathbf{f}_* \mid \mathbf{X}_f^*, \mathbf{X}, \mathbf{y}, \boldsymbol{\xi}_{\delta})$ see Appendix \ref{app:PE_disc}.

%--------------------------------------------------------------------------------------------------
% 2.3
\subsection{Physics-informed priors for biased data acquisition }\label{sec:PI_biased_sensors}

We now consider the setting where the physical model is perfect, but the observation errors are dependent.  For simplicity suppose that only the data for the function $u(x),$ $\mathbf{y}_u$ is biased. A model for this situation can be set up as 

\begin{equation*} 
  \begin{split}
    y_u & = u(x_u) + \text{Bias}(x_u) + \varepsilon_u, \text{ where } \text{Bias}(x) \sim GP(0, K_{\text{Bias}_u}(x,x'))\\
	y_f &= f(x_f) + \varepsilon_f.
  \end{split}
\end{equation*}
Mathematically the model formulation is similar to the model in Section \ref{sec:BCPI_main} with the difference that the discrepancy kernel, $\mathbf{K}_\delta$ is replaced by the Bias kernel $\mathbf{K}_{\text{Bias}},$ but these have the same prior formulation. Hence, the differences are in the interpretation of the discrepancy/bias term and its consequences for predictions.
Here we want to account for bias in the observed data and then remove the bias in the model predictions. 
The physics-informed prior is identical to Section \ref{sec:BCPI_main}, and the vector of the parameters $\boldsymbol{\xi}$ has been augmented with the vector $\boldsymbol{\theta}_{\text{B}},$ which is the Bias GP hyperparameters, and we denote the kernel parameters 
$\boldsymbol{\xi}_\text{Bias} = (\boldsymbol \theta, \boldsymbol{\theta}_B,  \boldsymbol \phi, \sigma_u, \sigma_f).$ In order to make predictions at new points $X^*_u,$ $u(X^*_u) = \mathbf{u}^*$ we have that the conditional distribution $p(\mathbf{u}^* \mid \mathbf{X}^*_u, \mathbf{X}, \mathbf{y}, \boldsymbol{\xi}_{\text{Bias}})$ is multivariate Gaussian (see Appendix \ref{app:Biased_meas}) and more specifically 

\begin{align*}
p(\mathbf{u}_* \mid \mathbf{X}_u^*, \mathbf{X}, \mathbf{y}, \boldsymbol{\xi}_\text{Bias}) &= \mathcal{N}(\boldsymbol{\mu}_u^*, \boldsymbol{\Sigma}_u^*) \\
\boldsymbol{\mu_u^*} &= \mu_u\mathbf{(X_u^*)}+ \mathbf{V}_u^*{^T} (\mathbf{K_{\text{Bias}}}+\mathbf{S})^{-1} (\mathbf{y}-\boldsymbol{\mu})\\
\boldsymbol{\Sigma_u^*} &= K_{uu}(\mathbf{X}_u^*,\mathbf{X}_u^*) - \mathbf{V}_u^*{^T} (\mathbf{K_{\text{Bias}}}+\mathbf{S})^{-1} \mathbf{V}_u^*,
\end{align*}
where $\mathbf{V}_u^*{^T} = \begin{bmatrix}  K_{uu}(\mathbf{X}_u^*,\mathbf{X}_u) &  K_{uf}(\mathbf{X}_u^*,\mathbf{X}_f) \end{bmatrix}.$ In contrast to Section \ref{sec:BCPI_main} where we learn the missing physics and this helps to improve model predictions, we now remove the bias in the predictions. \\
The conditional distribution $p(\mathbf{f}_* \mid \mathbf{X}_f^*, \mathbf{X}, \mathbf{y}, \boldsymbol{\xi})$ is multivariate Gaussian and more specifically

\begin{align*}
p(\mathbf{f}_* \mid \mathbf{X}_f^*, \mathbf{X}, \mathbf{y}, \boldsymbol{\xi}_B) &= \mathcal{N}(\boldsymbol{\mu}_f^*, \boldsymbol{\Sigma}_f^*) \\
\boldsymbol{\mu_f^*} &= \mu_f\mathbf{(X_f^*)}+ \mathbf{V}_f^*{^T} (\mathbf{K_{\text{Bias}}}+\mathbf{S})^{-1} (\mathbf{y}-\boldsymbol{\mu})\\
\boldsymbol{\Sigma_f^*} &= K_{ff}(\mathbf{X}_f^*,\mathbf{X}_f^*) - \mathbf{V}_f^*{^T} (\mathbf{K_{\text{Bias}}}+\mathbf{S})^{-1} \mathbf{V}_f^*,
\end{align*}
where $\mathbf{V}_f^*{^T} = \begin{bmatrix}  K_{fu}(\mathbf{X}_f^*,\mathbf{X}_u) &  K_{ff}(\mathbf{X}_f^*,\mathbf{X}_f) \end{bmatrix}.$ 

% -----------------------------------------------------------------------
% -----------------------------------------------------------------------
% -----------------------------------------------------------------------

\section{Synthetic Case Studies with Windkessel (WK) models}\label{sec:WK_synth}
In this section, we present a case study where the real physical process is more complex
than our modelling choice. More specifically, we  use the arterial Windkessel models which are
deterministic physical models describing the hemodynamics of the heart. First, we  consider
a synthetic case study where we use noisy simulated data from the physical model. Our goal is
to use the fully Bayesian physics-informed prior in order to infer and quantify the uncertainty of
the physical and noise parameters but also to generate model predictions. In a second synthetic
case study, we  simulate data from a more complex physical model than our modelling choice.
These models have mathematical connections that are described in Section \ref{sec:WK}. Our goal is to
infer the parameters of the more complex model by incorporating in the physics-informed prior a
discrepancy function. We also demonstrate the flexibility of this approach by considering different
kernel functions.

\subsection{Windkessel models} \label{sec:WK}
\begin{figure}[h!]
	\begin{minipage}[c]{0.5\textwidth}
		\includegraphics[width=\textwidth]{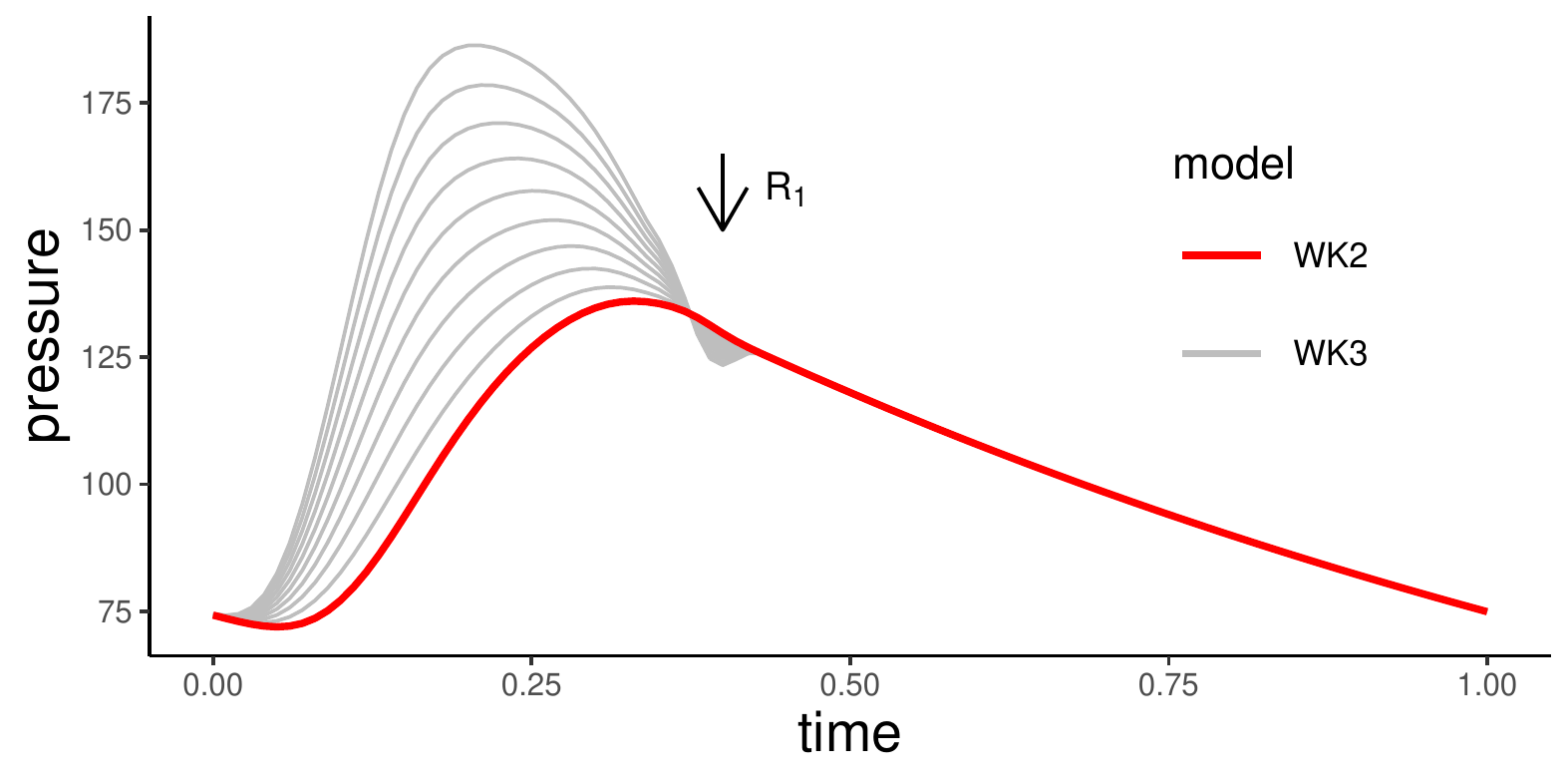} 
	\end{minipage}\hfill
	\begin{minipage}[c]{0.45\textwidth}
	   \caption{Blood pressure generated from WK2 model (red) and for a range of range of $R_1$ values $[0.01,0.2]$ from WK3 model (grey). The inflow and $C$ values are identical for both models. The amplitude of WK3-generated curve decreases linearly with $R_1,$ while the models become equivalent for $R_1 = 0.$} \label{fig:WK23}
	\end{minipage}
\end{figure}

The arterial Windkessel models \citep{westerhof2009arterial} describe the hemodymanics of the heart in terms of physically interpretable parameters. The simplest model, the Windkessel 2 parameters model (WK2) describes the relationship between blood pressure, $P(t)$ and blood inflow, $Q(t)$ by two key physical parameters, the total vascular resistance $R$ and arterial compliance, $C$ and it is defined by the following linear differential equation
\begin{equation}
	 Q(t) = \frac{1}{R}P(t) + C \frac{dP(t)}{dt} \label{eq:WK2}. 
\end{equation}
This model is the basis for building more complex physical models. For example, the Windkessel 3 (WK3) parameters model introduce a second resistive parameter $R_1$ and is given by the following linear differential equation
\begin{equation}
	 \frac{d P(t)}{d t} + \frac{P(t)}{R_2C} = \frac{Q(t)}{C} \left (1 + \frac{ R_1}{R_2} \right ) + R_1 \frac{d Q(t)}{dt}.
	 \label{eq:WK3}
\end{equation}
The inclusion of the third parameter increases flexibility and might improve fitting to the observed data. However, it overestimates the total arterial compliance, $C$ \citep{segers2008three}. In Figure \ref{fig:WK23}, we see the blood pressure waveform for the WK2 model (red) and for a range of $R_1$ values of the WK3 model (grey). From a modelling perspective, the $R_1$ parameter controls the discrepancy between the two models. An important connection for the synthetic case study is that the ratio of mean pressure over inflow equals $R$ in the WK2 model, while for the WK3 model this ratio equal to $R_1+R_2$ \citep{westerhof2009arterial}.

\subsection{WK Case Study 1: Full Bayesian analyses}\label{sec:Unbiased_WK2}
In this study, we simulate noisy data from the deterministic WK2 model \eqref{eq:WK2} and use a physics-informed probabilistic WK2 model to estimate the physical parameters and quantify the uncertainty. We also produce model predictions for blood pressure, $P(t)$ and blood inflow, $Q(t).$ To demonstrate this approach's flexibility and do a sensitivity analysis of the GP prior choice, we also consider three kernels. The squared exponential (SE), the rational quadratic (RQ) and the periodic kernel (Per). The periodic kernel is a natural choice as the blood pressure is a periodic phenomenon that repeats at each cardiac cycle (time between two consecutive heartbeats).
	
\begin{figure}[h!]
	%\centering
	\includegraphics[width=1\textwidth,height=0.35\textheight]{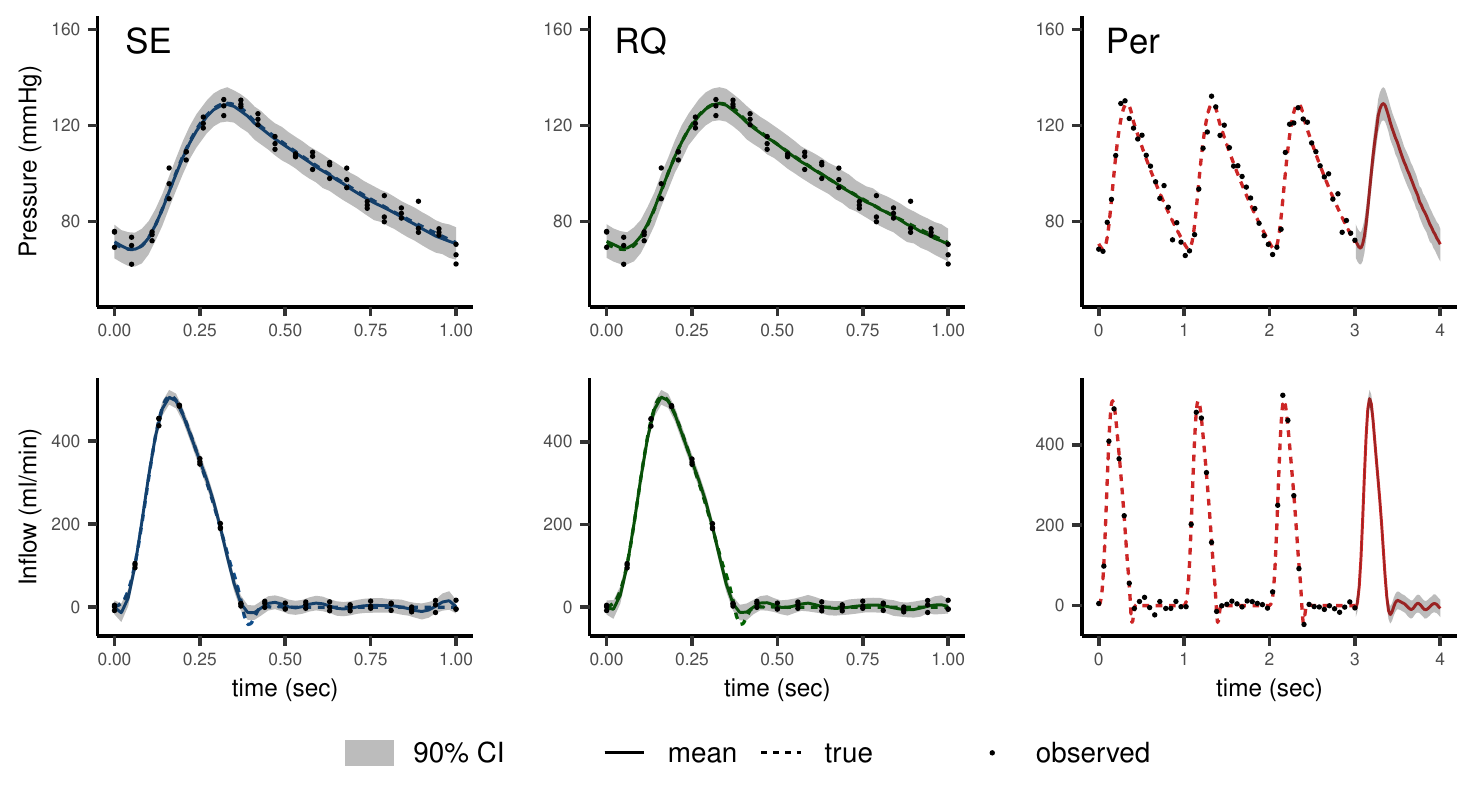}
	\vspace*{-4mm}	
	\caption{Predictions for all kernels. SE (squared exponential), RQ (rational quadratic) and Per (periodic).}
	\label{fig:Pred_all_kernels_Unbiased}
\end{figure}

To simulate blood pressure data, $P(t)$ from the deterministic WK2 model, we choose a given observed blood inflow, $Q(t)$ (see Figure \ref{fig:Pred_all_kernels_Unbiased}, bottom) and set parameter values, $R=1, C=1.1.$  Gaussian i.i.d. noise is added to both pressure and inflow as follows, $y_P = P(t)+\varepsilon_P$ and $y_Q=Q(t)+\varepsilon_Q,$ where $\varepsilon_P \sim N(0,4^2)$ and $\varepsilon_Q \sim N(0,10^2).$ We simulated replicates at each observed temporal location, $t_i$ by synchronizing three blood pressure cycles in one (see Figure \ref{fig:Pred_all_kernels_Unbiased}, where the third column of plots is the unsynchronized data and in the first column is the synchronized) replicates are used as this helps to separate the signal from noise.
	
We construct physics-informed prior for the WK2 model by assuming a GP prior on pressure, $P^{\text{WK2}}(t_P)\sim GP(\mu_P, K_{PP}(t_P, t_P')).$ Three models with different covariance functions (squared exponential, rational quadratic and periodic) are considered. For all three models we assume a constant mean $\mu$ and the WK2 physics-informed prior is defined as follows
\begin{equation} 
	\begin{split}
	  y_P & = P^{\text{WK2}}(t_P) + \varepsilon_P\\
	  y_Q & = Q^{\text{WK2}}(t_Q) + \varepsilon_Q, 
	\end{split}
\end{equation}
where $P^{\text{WK2}}(t_P) \sim GP(\mu_P, K(t_P, t_P')), \varepsilon_P \sim N(0, \sigma^2_P)$ and $\varepsilon_Q \sim N(0,\sigma^2_Q).$ This results in the following multi-output GP prior
\begin{equation}
p(\mathbf{y}\mid \boldsymbol \theta,  \boldsymbol \phi, \sigma_P, \sigma_Q) = \mathcal{N}(\boldsymbol{\mu}, \mathbf{K}),
\end{equation}
where \\
$
\bf{y} = \begin{bmatrix}  \bf{y}_P \\  \bf{y}_Q \end{bmatrix},
$ 
$\boldsymbol{\mu} = \begin{bmatrix}  \boldsymbol{\mu}_P \\   R^{-1}\boldsymbol{\mu}_P \end{bmatrix} $ and 
$\\
\mathbf{K} =
\begin{bmatrix}
K_{PP}(\mathbf{t}_P, \mathbf{t}_P \mid \boldsymbol \theta) + \sigma_P^2 I_P & K_{PQ}(\mathbf{t}_P, \mathbf{t}_Q \mid \boldsymbol \theta,  \boldsymbol \phi)\\
K_{QP}(\mathbf{t}_Q, \mathbf{t}_P \mid \boldsymbol \theta,  \boldsymbol \phi) & K_{QQ}(\mathbf{t}_Q, \mathbf{t}_Q \mid \boldsymbol \theta,  \boldsymbol \phi) + \sigma_Q^2 I_Q
\end{bmatrix} 
$ (see Appendix \ref{app:WK} for more details on the elements of the matrix $\mathbf{K}$).	

\vspace{0.2cm}
\noindent Furthermore, uniform priors are assigned on the physical parameters of interest on a range of reasonable values, 
$R,C \sim \mathcal{U}(0.5,3)$ and also weakly informative priors to the other model hyperparameters (see Appendix \ref{app:WK}, WK2 model). 
	
\begin{figure}[h!]
	%\centering
	\includegraphics[width=1\textwidth,height=0.2\textheight]{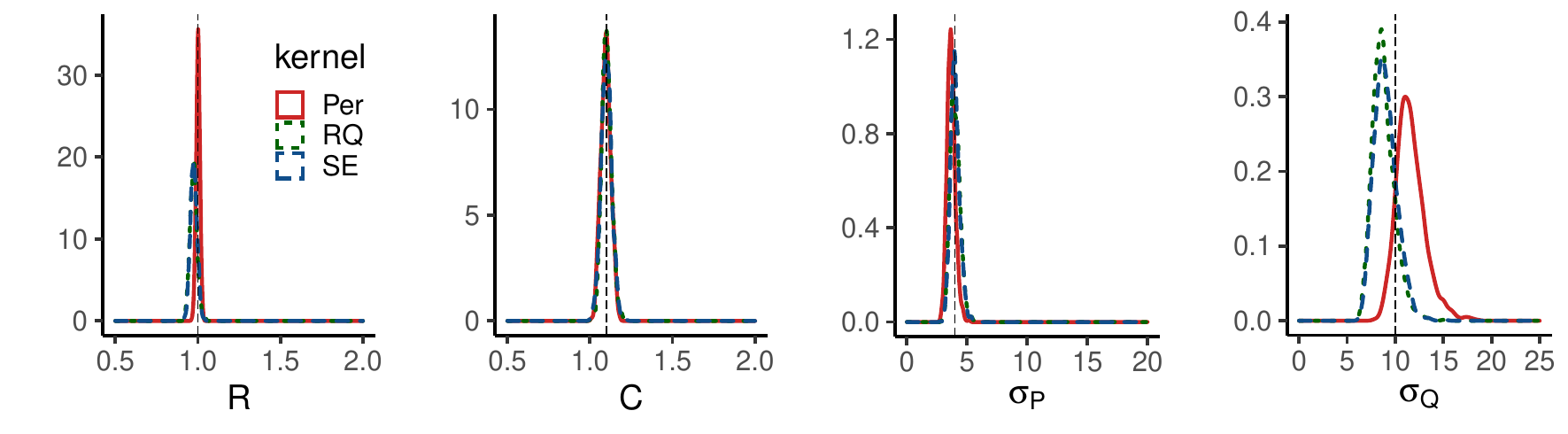}
	\vspace*{-4mm}	
	\caption{Posterior distributions of the physical and noise parameters for the models: SE (squared exponential), RQ (rational quadratic) and Per (periodic).}
	\label{fig:Post_all_kernels_Unbiased}
\end{figure}
We fit the fully Bayesian physics-informed WK2 model to the observed data. In Figure \ref{fig:Post_all_kernels_Unbiased}, the posterior distributions for the physical and noise parameters are plotted. We see that all three models estimate the resistance value, $R$ accurately , and the uncertainty is relatively small. The posteriors for the squared exponential (SE) and the rational quadratic (RQ) kernels are identical, while for the periodic kernel the uncertainty is slightly reduced, and this is probably because we impose more prior information by encoding on the model that the phenomenon repeats itself exactly after some length $p$ (here $p=1$ sec., see Figure \ref{fig:Post_all_kernels_Unbiased}, right). The posterior of the compliance parameter, $C,$ is concentrated around the true value with small uncertainty and is also identical for all three models. All models also estimate the pressure noise, $\sigma_P$ well. The difference is found for the posterior of the blood inflow noise parameter, $\sigma_Q.$ In Figure \ref{fig:Heat_Unbiased_pred}, bottom plots, we see that the inflow is constant two-thirds of the time and equals 0 (this happens during diastole, where the aortic valve is closed, and consequently, the blood inflow is 0). Therefore, it is harder for the models to smooth the observed data. However, this is an advantage of taking a fully Bayesian approach since the true value of the noise parameter is within the $90\%$ credible intervals. In Figure \ref{fig:Pred_all_kernels_Unbiased}, we see that all three models predict well blood pressure and blood inflow with relatively small uncertainty.
	
%----------------------------------	

\subsection{WK Case Study 2: Model discrepancy}\label{sec:Biased_model}
In this synthetic case study, the ground truth is a more complex model than our modelling choice. We simulate noisy data from the WK3 model and use the WK2 model as our modelling choice. More specifically, for a given inflow $Q(t),$ we simulate data from the deterministic WK3 model, $P(t)=P_{\text{WK3}}(Q(t), R_1=0.05, R_2=1, C=1.1).$  To create the observed pressure, $y_P$ and inflow $y_Q$ data, we add i.i.d. Gaussian noise as follows, $y_P = P(t)+\varepsilon_P,$ where $\varepsilon_P\sim N(0,4^2)$ and $y_Q = Q(t)+\varepsilon_Q,$ where $\varepsilon_Q\sim N(0,10^2).$
As described in Section \ref{sec:WK} we expect that $R^{\text{WK2}} = R_1^{\text{WK3}}+R_2^{\text{WK3}}$ and $C^{\text{WK2}}=C^{\text{WK3}}$ when the WK2 model is fitted to the WK3 data.
\begin{figure}[h!]
	%\centering
	\includegraphics[width=1\textwidth,height=0.4\textheight]{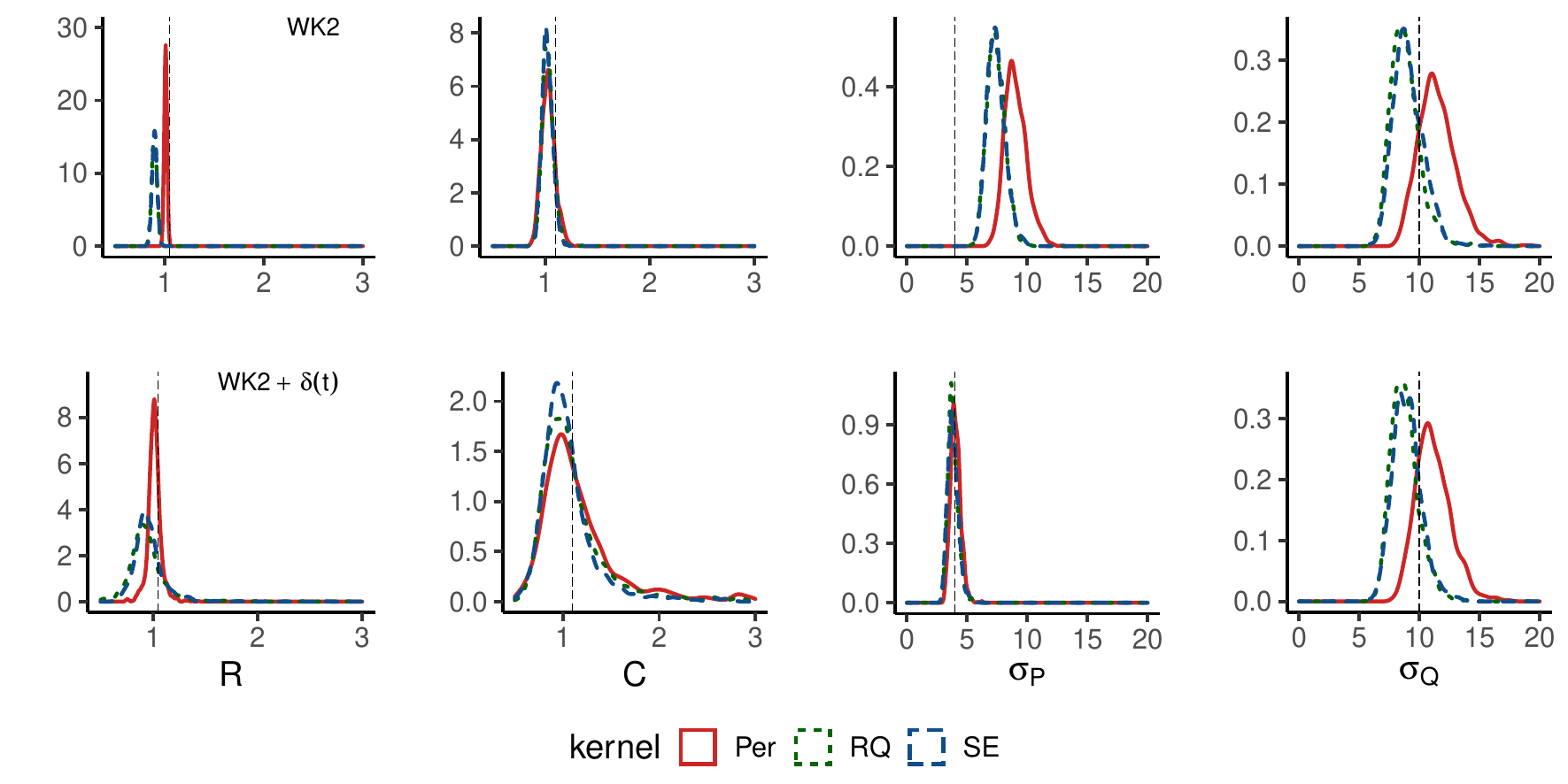}
	\vspace*{-4mm}	
	\caption{Posterior distributions for all kernels denoted as SE (squared exponential), RQ (rational quadratic) and Per (periodic) in a single plot (see colours). The first row of plots is the resulting posteriors if not accounting for model discrepancy (WK2) and the second row shows the posteriors of the model accounting for functional model discrepancy ($\text{WK2} + \delta(t)$).}
	\label{fig:Post_all_kernels_Biased}
\end{figure}
We consider two probabilistic models. The first model does not account for model discrepancy and it is identical to the model in Section \ref{sec:Unbiased_WK2}. The second model incorporates a functional discrepancy $\delta(t)$ in the physics-informed prior formulation and is defined as follows
\begin{equation} 
	\begin{split}
	  y_P & = P^{\text{WK2}}(t_P) + \delta(t_P) + \varepsilon_P\\
	  y_Q & = Q^{\text{WK2}}(t_Q) + \varepsilon_Q, 
	\end{split}
\end{equation}
where $P^{\text{WK2}}(t_P) \sim GP(\mu_P, K(t_P, t_P')), \varepsilon_P \sim N(0, \sigma^2_P)$ and $\varepsilon_Q \sim N(0,\sigma^2_Q)$ as in Section \ref{sec:Unbiased_WK2}. In addition we assume a GP prior for the model discrepancy $\delta(t_P),$ $\delta(t_P) \sim GP(\mu_P, K_{\delta}(t_P, t_P')),$ resulting in  the following multi-output GP prior
\begin{equation}
p(\mathbf{y}\mid \boldsymbol \theta, \boldsymbol{\theta}_\delta, \boldsymbol \phi, \sigma_P, \sigma_Q) = \mathcal{N}(\boldsymbol{\mu}, \mathbf{K}) 
\end{equation}
where 
$
\bf{y} = \begin{bmatrix}  \bf{y}_P \\  \bf{y}_Q \end{bmatrix},
$ 
$\boldsymbol{\mu} = \begin{bmatrix}  \boldsymbol{\mu}_P \\   R^{-1}\boldsymbol{\mu}_P \end{bmatrix} $ and \\
$
\mathbf{K} =
\begin{bmatrix}
K_{PP}(\mathbf{t}_P, \mathbf{t}_P \mid \boldsymbol \theta)+K_{\delta}(\mathbf{t}_P,\mathbf{t}_P\mid \boldsymbol \theta_{\delta}) + \sigma_P^2 I_P & K_{PQ}(\mathbf{t}_P, \mathbf{t}_Q \mid \boldsymbol \theta,  \boldsymbol \phi)\\
K_{QP}(\mathbf{t}_Q, \mathbf{t}_P \mid \boldsymbol \theta,  \boldsymbol \phi) & K_{QQ}(\mathbf{t}_Q, \mathbf{t}_Q \mid \boldsymbol \theta,  \boldsymbol \phi) + \sigma_Q^2 I_Q
\end{bmatrix}. 
$

\vspace{0.1 cm}

\noindent As in the unbiased physical model case study, we assign uniform priors on the physical parameters of interest, 
$R,C \sim \mathcal{U}(0.5,3)$ and weakly informative priors to the other model hyperparameters (see Appendix \ref{app:WK}, $\text{WK2}+\delta(t)$ model). Finally, for the models with squared exponential (SE) and rational quadratic (RQ) kernels, we use the squared exponential kernel for $\delta(t),$ while for the model with the periodic (Per) kernel we use a periodic kernel for $\delta(t)$ as well.

\begin{figure}[h!]
	%\centering
	\includegraphics[width=1\textwidth,height=0.4\textheight]{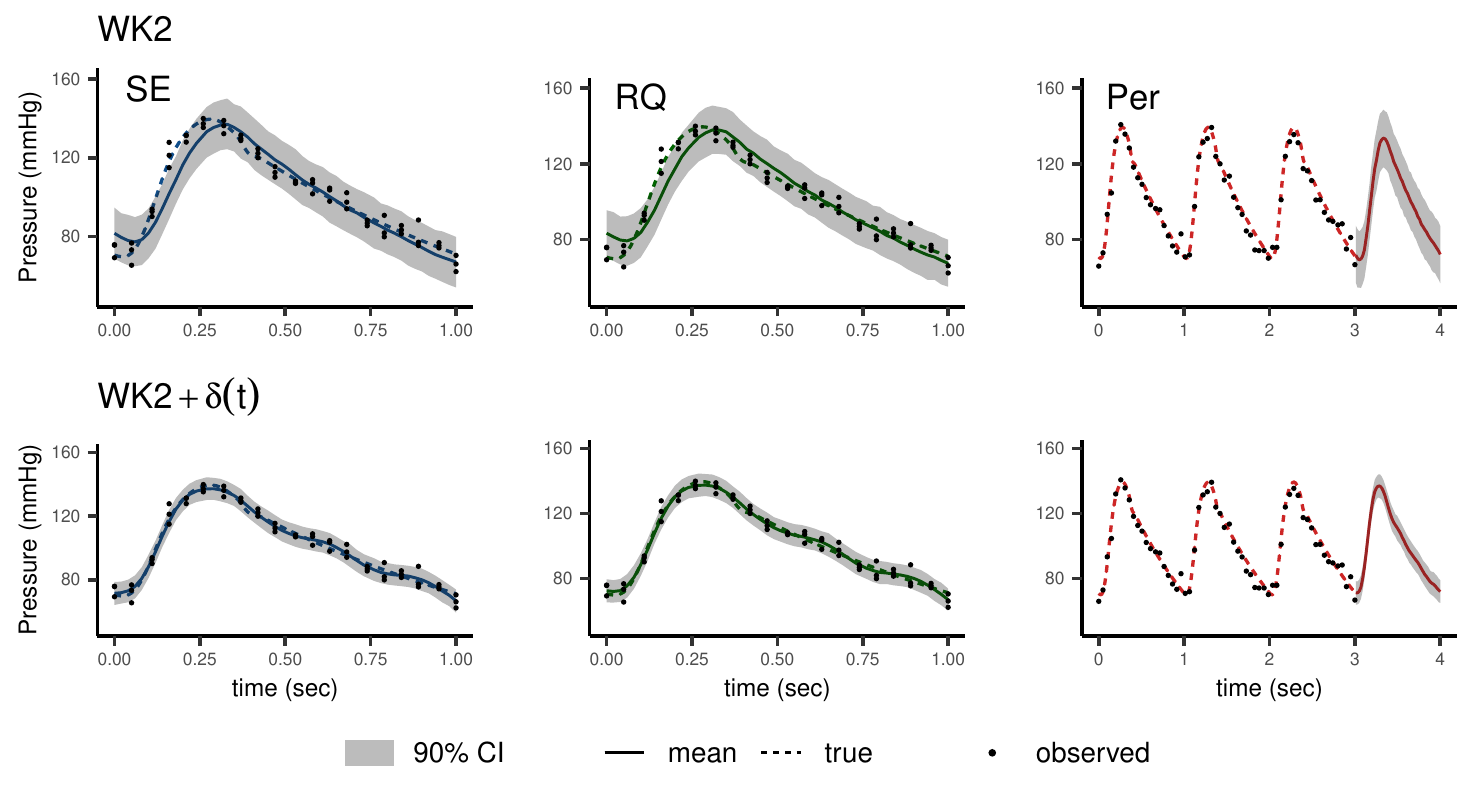}
	\vspace*{-4mm}	
	\caption{Blood pressure predictions for all kernels denoted as SE (squared exponential), RQ (rational quadratic) and Per (periodic). The first row of plots is the model without accounting for discrepancy (WK2) and the second row the model accounting for model discrepancy.}
	\label{fig:Pred_Pres_all_kernels_Biased}
\end{figure}

\begin{figure}[h!]
	%\centering
	\includegraphics[width=1\textwidth,height=0.4\textheight]{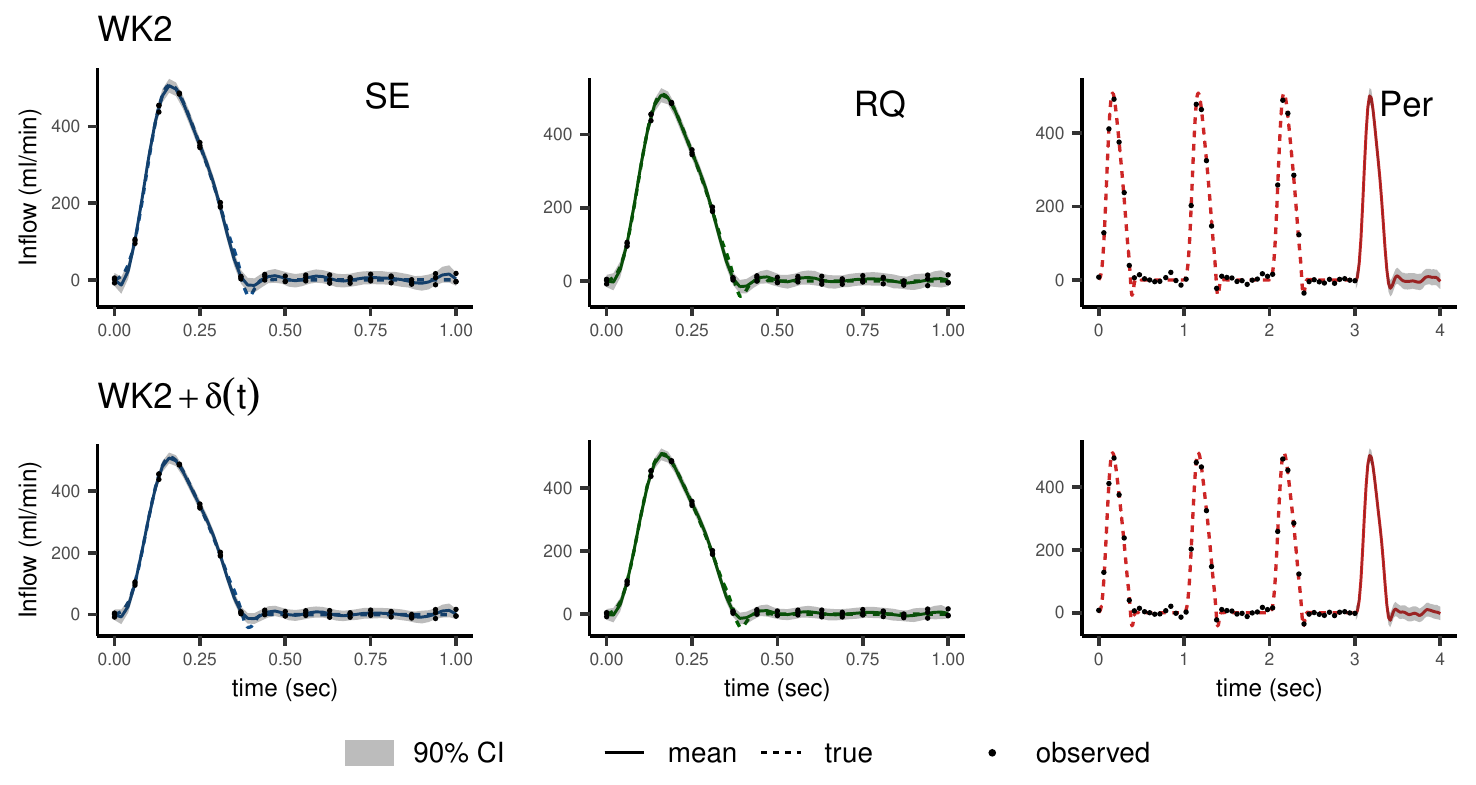}
	\vspace*{-4mm}	
	\caption{Blood inflow predictions for all kernels denoted as SE (squared exponential), RQ (rational quadratic) and Per (periodic). The first row of plots is the model without accounting for discrepancy (WK2) and the second row the model accounting for model discrepancy.}
	\label{fig:Pred_Flow_all_kernels_Biased}
\end{figure}

Models not accounting and accounting for model discrepancy  are fitted. Results are found in Figures \ref{fig:Post_all_kernels_Biased}, \ref{fig:Pred_Pres_all_kernels_Biased} and \ref{fig:Pred_Flow_all_kernels_Biased}. Results for models not accounting for discrepancy are in upper rows, and corresponding models accounting for model discrepancy in lower rows.  In upper row in Figure \ref{fig:Post_all_kernels_Biased} we find that if we do not account for model discrepancy the posteriors of the physical parameters ($R$ and $C$) are biased and overconfident. In particular, the resistance parameter $R$  is underestimated for the square exponential (SE) and rational quadratic (RQ) covariance models, while for the periodic (Per) model, the uncertainty is very small. For the compliance parameter, $C,$ the posterior distributions of all three models are almost identical with a relatively small uncertainty, and the true value is at the tail of the posteriors. The physics-informed WK2 models can not capture the observed blood data well (see Figure \ref{fig:Pred_Pres_all_kernels_Biased}), resulting in overestimating the noise parameter $\sigma_P$ (see Figure \ref{fig:Post_all_kernels_Biased}). 

Accounting for model discrepancy ($\text{WK2}+ \delta(t)$ in Figures \ref{fig:Post_all_kernels_Biased}, \ref{fig:Pred_Pres_all_kernels_Biased} and \ref{fig:Pred_Flow_all_kernels_Biased}) results in a more reasonable quantification of the uncertainty in the physical parameters ($R$ and $C$, Figure \ref{fig:Post_all_kernels_Biased}, bottom row plots). The parameter uncertainties have now increased, which is sensible given that the WK2 model is a simplification of the real data generating process. However, now it covers the true resistance value R. This also holds for the compliance parameter $C.$ Further, the noise parameter $\sigma_P$ is estimated accurately, and this means that the model has learned the discrepancy between the two models. We also produce blood pressure and blood inflow predictions. In Figure \ref{fig:Pred_Pres_all_kernels_Biased}, we see that if we do not account for model discrepancy (WK2), the models (SE, RQ and Per) can not fit the observed data well the prediction uncertainty is quite large. By accounting for model discrepancy ($\text{WK2}+\delta(t)$ model), the probabilistic model has learned the missing physics and this has significantly reduced the uncertainty in the model predictions. In Figure \ref{fig:Pred_Flow_all_kernels_Biased}, we see that for both the WK2 and the $\text{WK2} + \delta(t)$ models, the predictions are accurate with small uncertainty since there is not any discrepancy in the data generating process for the blood inflow.

\begin{figure}[h!]
	%\centering
	\includegraphics[width=1\textwidth,height=0.5\textheight]{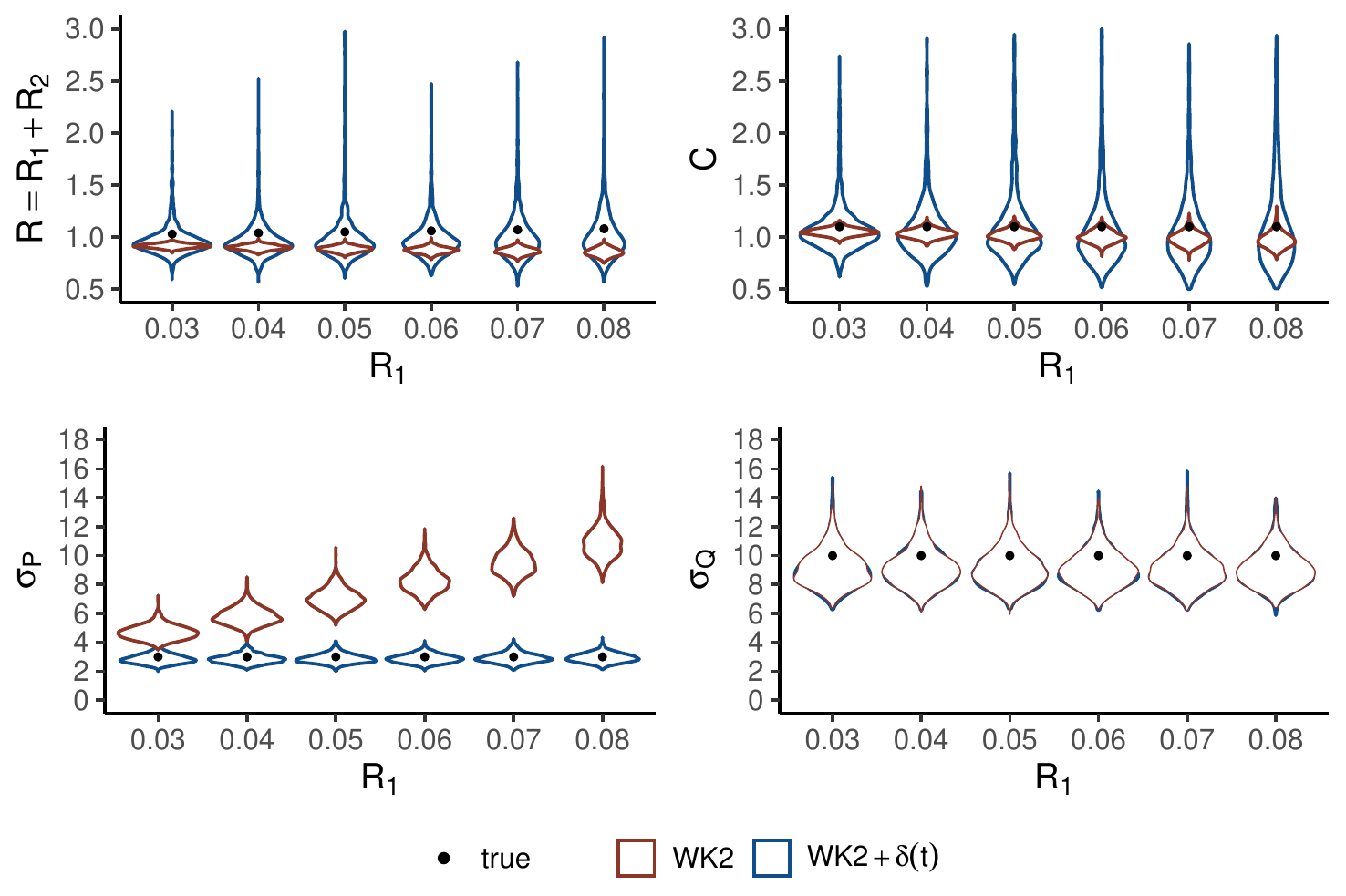}
	\vspace*{-4mm}	
	\caption{Posterior distributions for a range of $R_1$ values. Larger $R_1$ values result to larger discrepancy between the deterministic WK3 (true) and WK2 (modelling choice) models. For both probabilistic models (WK2 and $\text{WK2} + \delta(t)$) the squared exponential kernel is used.}
	\label{fig:range_R1}
\end{figure}

As in Section \ref{sec:WK}, the $R_1$ parameter controls the discrepancy between the WK3 and WK2 models. We simulate noisy data from the WK3 model again, but now for a range of $R_1$ values, $R_1=0.03,\dots , 0.08,$ and we fit the two models (WK2 and $\text{WK2}+\delta(t)$) again in order to obtain posterior distribution of the physical and noise parameters. In Figure \ref{fig:range_R1}, top left plot, we observe that with increasing discrepancy (corresponds to increased $R_1$ value) the bias of the resistance parameter $R$ for the WK2 model (red posteriors) increases as well while accounting for model discrepancy (blue posteriors) produces reasonable quantification of the uncertainty for all $R_1$ values. In the top-right plot of Figure \ref{fig:range_R1}, we observe that for the WK2 model (red posteriors), the true value of the compliance parameter $C$ is at the tail of the posterior in all cases while accounting for model discrepancy (blue posteriors) produces reasonable quantification of the uncertainty and the posterior covers the true value again. In the bottom left plot, we see that the overestimation of $\sigma_P$ increases with the discrepancy between the two models. While accounting for discrepancy, the model estimates the pressure noise parameter $\sigma_P$ accurately. Furthermore, in the bottom right plot, we see that noise estimates for both models are identical since the blood inflow has no discrepancy. 

%---------------------------------------------

\section{Synthetic Case Studies: Heat Flow} \label{sec:HF}
In this section, we consider the Heat equation which is one of the most important differential equations in science and engineering. To demonstrate our approach, we use only one spatial dimension. %However, the generalization to the three-dimensional space is straightforward. 
First, we  briefly describe the physical model and its physics-informed prior and then consider two synthetic case studies. In the first study, we simulate data from the model and add i.i.d. Gaussian noise. In the second study, we assume that the data acquisition process is biased and that this bias can be described by a non-linear function. Our goal is to estimate the model’s physical and noise parameters and quantify their uncertainty.  We also produce model predictions.
	
\subsection{Heat equation}
The non-homogeneous Heat equation is given by the following space-time dependent differential equation
\begin{equation}
	\frac{\partial u}{\partial t} - \alpha \nabla^2 u = f,
\end{equation}
where $u$ describes the heat distribution in space and time and $f$ is the forcing (heat generation source). We treat the thermal conductivity parameter, $\alpha$ as unknown and we wish to infer its value using noisy observed data.
In the 1D case the heat equation describes the distribution of heat, $u(t,x)$ in a thin metal rod and the differential equation reduces to 
\begin{equation}
   \frac{\partial u(t,x)}{\partial t} - \alpha \frac{\partial^2 u(t,x)}{\partial^2 x} = f(t,x).
\end{equation}	
For $\alpha=1,$ the functions $f(t,x) = \exp(-t)(4\pi^2-1)\sin(2\pi x)$ and $u(t,x) = \exp(-t)\sin(2\pi x)$ satisfy this equation. This solution is used to simulate data for the synthetic case studies. 
%-----------------------------------------------------------------------------------

\subsection{HF Case Study 1: Fully Bayesian analysis}
\begin{figure}[h!]
	\centering
	\includegraphics[width=1\textwidth,height=0.3\textheight]{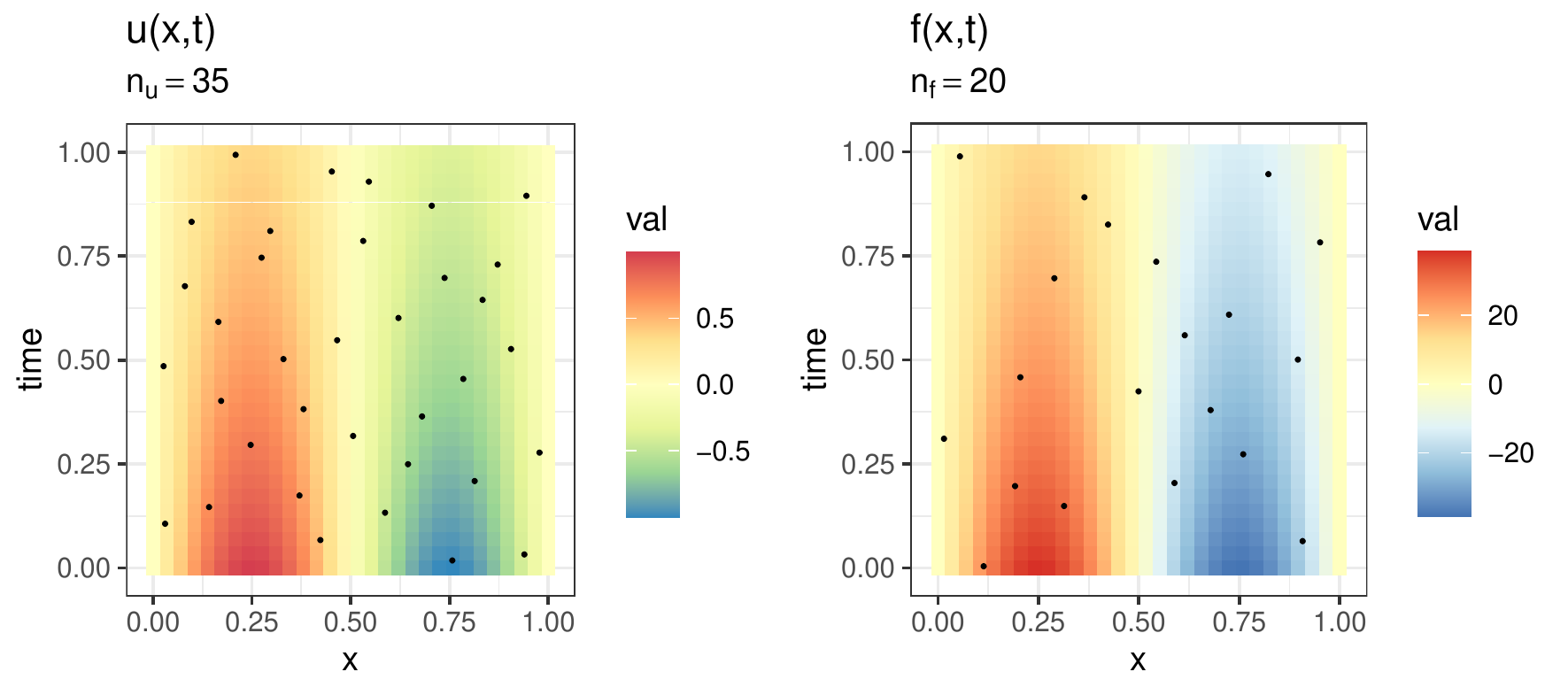}
	\vspace*{-4mm}	
	\caption{Heat, $u(t,x)$ and forcing, $f(t,x)$ data.}
	\label{fig:Heat_observed_uf}
\end{figure}

We simulate data from the model for $\alpha=1$ and the solution given in Section 4.1 and add i.i.d. noise. More specifically, we simulate 35 data points for $u(t,x)$ and 20 data points for $f(t,x)$ sampled randomly on $[0,1]^2$ (see Figure \ref{fig:Heat_observed_uf}). We add Gaussian noise to the simulated $u$ and $f$ values and we obtain the observed data as follows,$y_u = u(t,x) + \varepsilon_u,$ where $\varepsilon_u\sim N(0,0.2^2)$ and $y_f = f(t,x) + \varepsilon_f,$ where $\varepsilon_f\sim N(0,1^2).$

\begin{figure}[h!]
	\centering
	\includegraphics[width=1\textwidth,height=0.3\textheight]{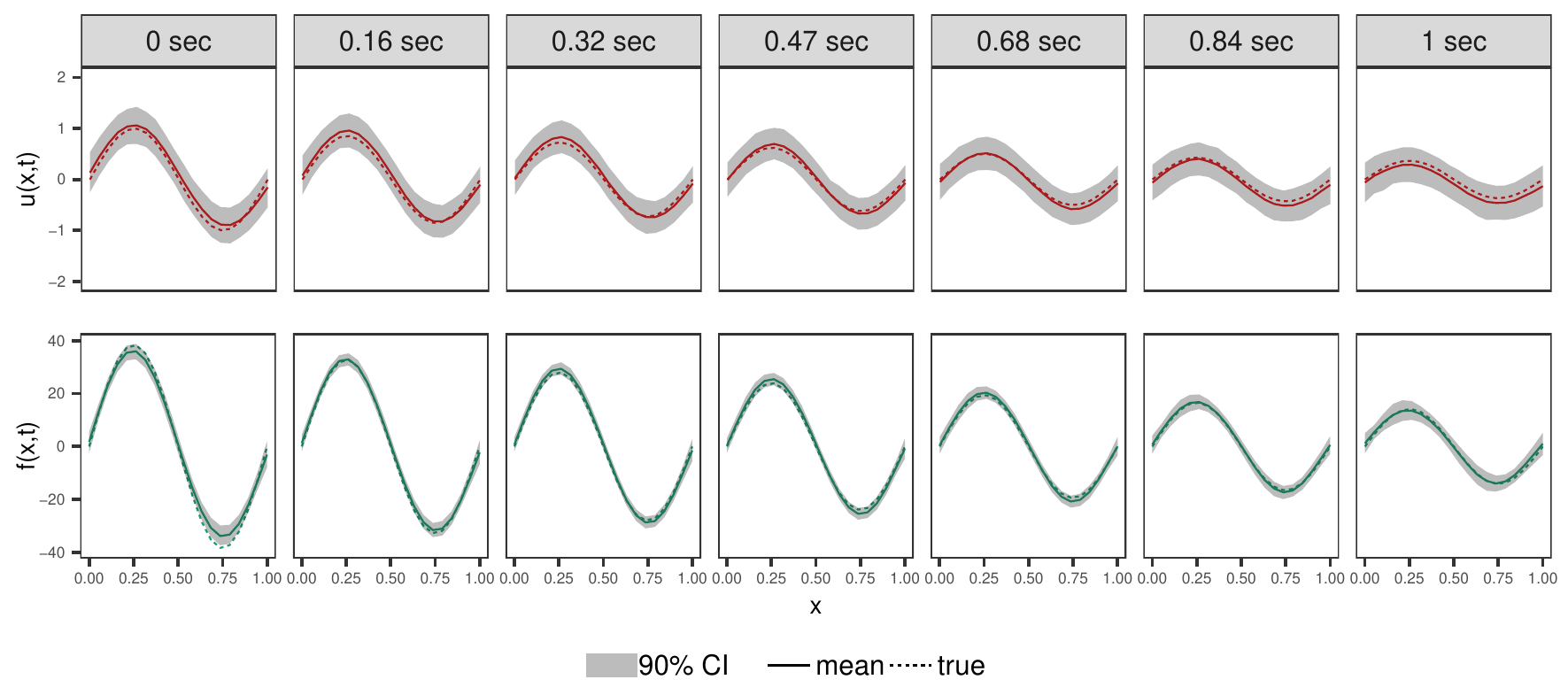}
	\vspace*{-4mm}	
	\caption{Predictions for unbiased sensor data, plotted as time evolution snap shots. The solid line represents the mean and the shaded region is the $90\%$ credible interval, while the dashed line is the true heat distribution.}
	\label{fig:Heat_Unbiased_pred}
\end{figure}

To develop the physics-informed prior for the Heat equation, we assume that the heat follows a GP prior, $u(t,x)\sim GP(\mu,K((t,x), (t',x')))$ where we use an anisotropic squared exponential kernel,

\begin{equation*}
    K_{uu}((t,x), (t',x')) = \sigma^2 exp\left(-\frac{1}{2l_t^2}(t-t')^2\right)exp\left(-\frac{1}{2l_x^2}(x-x')^2\right)
\end{equation*}
and $\mu$ is a constant. We derive the physics-informed prior, which is a  multi-output GP of $u(t,x)$ and $f(t,x)$ as detailed in the Section \ref{sec:method}. We use a uniform prior for $\alpha,$ $\alpha \sim U[0,10]$ and weakly informative priors for the hyperparameters of the physics-informed prior (see Appendix B.2, for details on the kernel hyperparameters and the physics-informed kernel).
\begin{figure}[h!]
	\centering
	\includegraphics[width=1\textwidth,height=0.15\textheight]{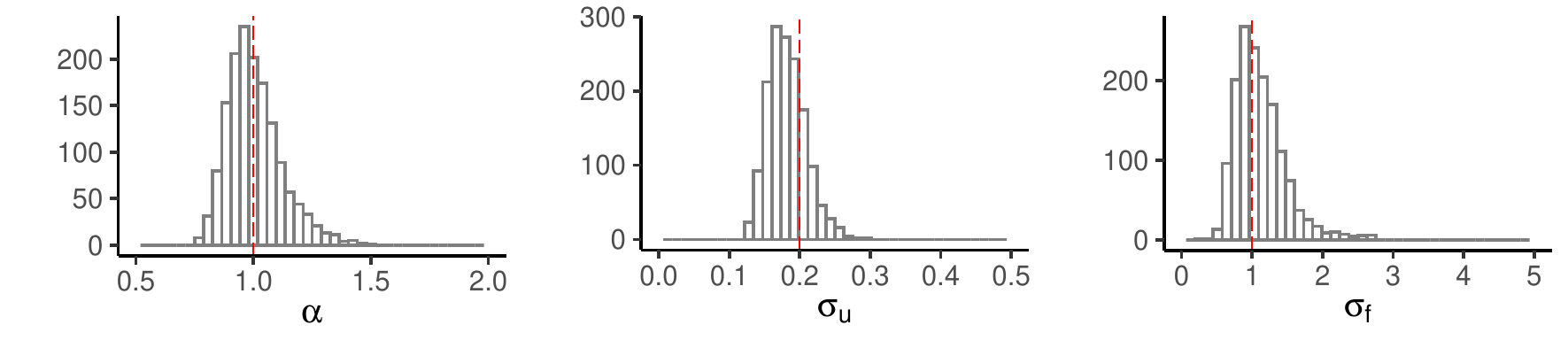}
	\vspace*{-4mm}	
	\caption{Posterior distributions for the parameters of interest ($\alpha$ is the diffusion parameter and $\sigma_u$ and $\sigma_f$ are the heat and forcing noise standard deviations respectively). The red dashed line is the true value.}
	\label{fig:Post_Heat_Unbiased}
\end{figure}
To infer the parameters, we use Hamiltonian Monte Carlo sampling. In Figure \ref{fig:Post_Heat_Unbiased}, we observe that for the physical parameter $\alpha,$ the posterior density is concentrated around the true value, and the uncertainty is relatively small. The same holds for the forcing noise estimation (Figure \ref{fig:Post_Heat_Unbiased}, right), while the heat noise is slightly underestimated (Figure \ref{fig:Post_Heat_Unbiased}, middle). However, the $90\%$ credible interval covers the true value and this is an advantage of the fully Bayesian approach. In Figure \ref{fig:Heat_Unbiased_pred}, we produce predictions for both $u$ and $f.$ We see that both prediction means are very accurate, and also the prediction uncertainty is small.
	
\subsection{HF Case Study 2: Biased sensor observations}
\begin{figure}[h!]
	\centering
	\includegraphics[width=1\textwidth,height=0.3\textheight]{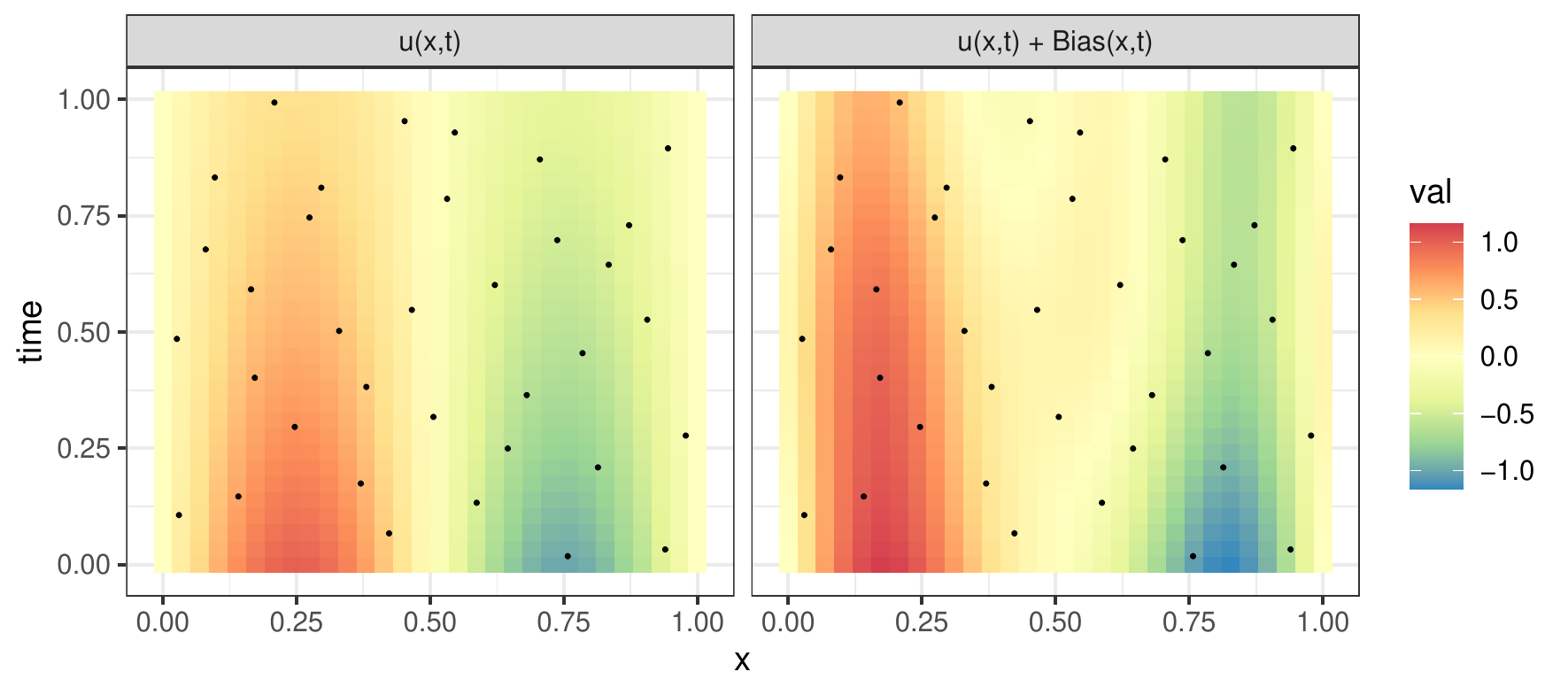}
	\vspace*{-4mm}	
	\caption{The left plot ($u(t,x)$) is the true heat distribution in space,x and time while the right plot ($u(x,t)+ \text{Bias(x,t)}$) is the heat distribution obtained from biased measurements. The black dots represent the observation points.}
	\label{fig:Heat_observed_Bias}
\end{figure}

For the hemodynamics models (introduces in Section \ref{sec:WK_synth}), we know that they are imperfect representations of the real process, and
thus it is reasonable to incorporate a discrepancy function in the model formulation. In contrast, we now assume that the heat equation can accurately describe the true process. However, the sensors that measure the heat, $u(t,x),$ create bias to the measurements, $y_u$ in a non-linear way. More specifically, to demonstrate a synthetic case study, we generate bias in the observational process by the following function,
$b(t,x) = \sin(4\pi x)/3+2t^2(1-t)^2$. We use the previously simulated data (unbiased sensor data), and we add bias according to this non-linear function. In Figure \ref{fig:Heat_observed_Bias}, we see that this function increases the absolute value of $u(x,t)$ towards the boundaries of the spatial domain and decreases the absolute value of $u(x,t)$ towards 0 in the middle. 

The approach now is similar to the approach where we considered a discrepancy function, but now this function is under the name Bias. The main reason for this is that this function does not learn the missing physics of the process. It is used as an auxiliary process, and it is removed when we use the model to predict. This also results in increased uncertainty in model predictions, as we will see shortly. 
	
\begin{figure}[h!]
	\centering
	\includegraphics[width=1\textwidth,height=0.3\textheight]{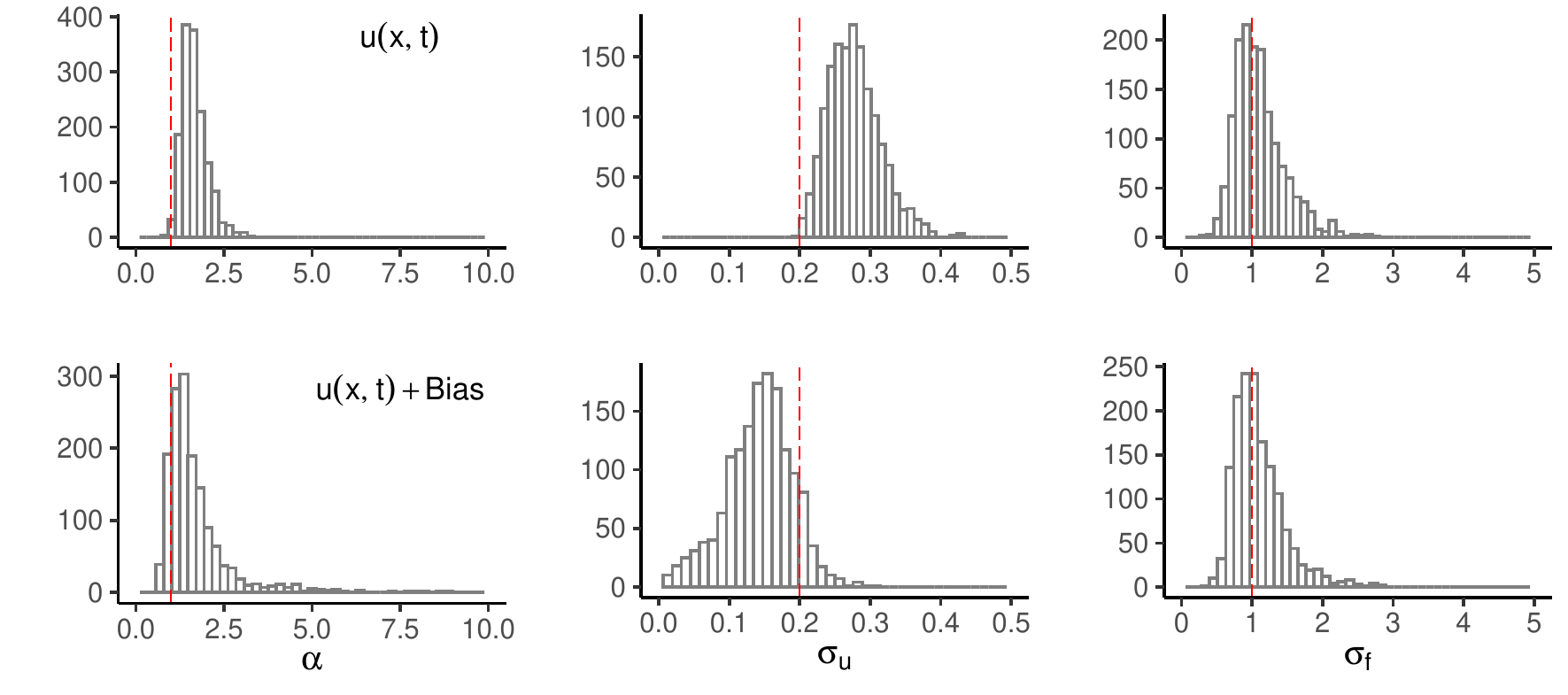}
	\vspace*{-4mm}	
	\caption{Posterior distributions for the parameters of interest ($\alpha$ is the diffusion parameter, $\sigma_u$ is the heat noise sd and $\sigma_f$ is the forcing noise sd). The red dashed line is the true values.}
		\label{fig:Post_Heat_Biased}
\end{figure}

\begin{figure}[h!]
   \centering
	\includegraphics[width=1\textwidth,height=0.35\textheight]{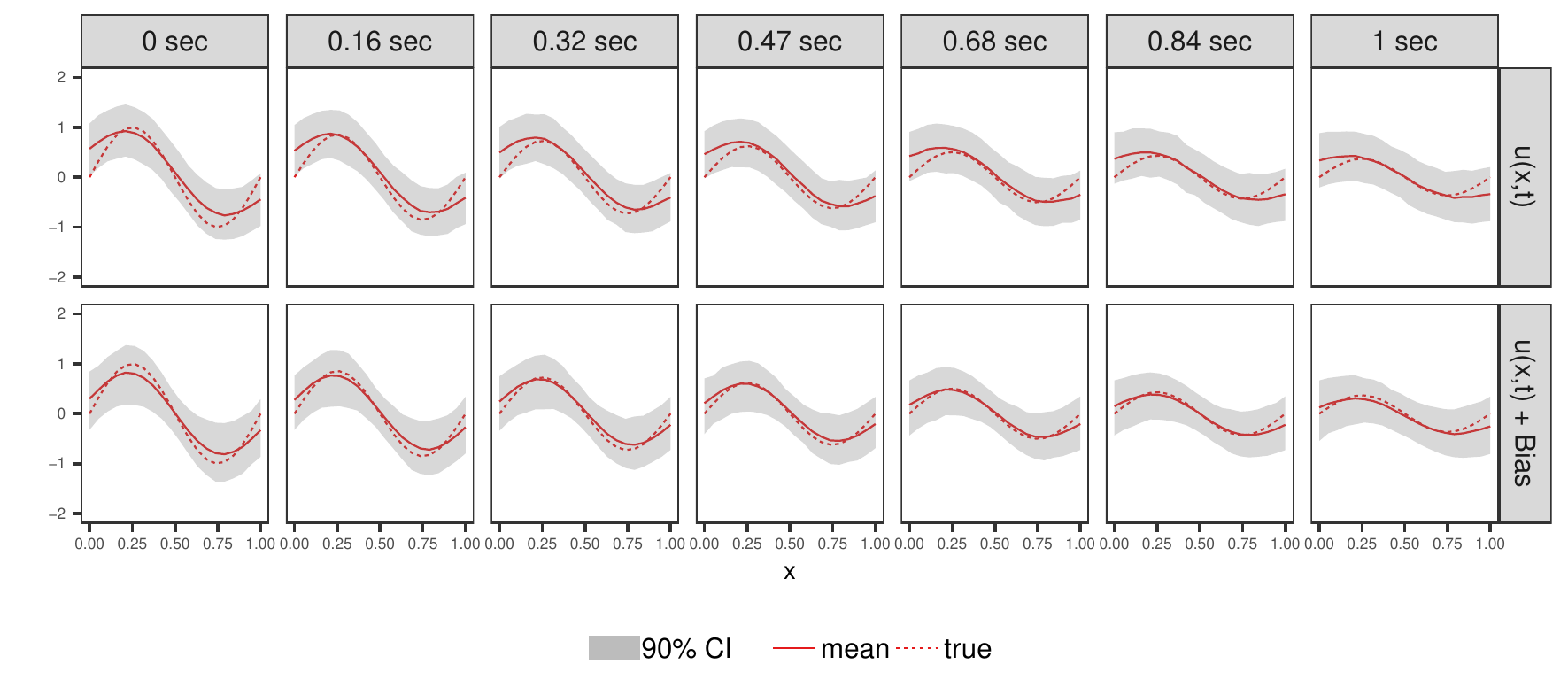}
	\vspace*{-4mm}	
	\caption{Predictions (time evolution snap shots) for biased sensor data. First row is the heat distribution at several temporal locations when not accounting for bias in the measurements while in the second row is the model which accounts for bias.}
	\label{fig:Heat_bias_pred}
\end{figure}

For the biased simulated data, we fit two models. The first model does not account for bias in the measurement process ($u(t,x)$ in Figures \ref{fig:Post_Heat_Biased} and \ref{fig:Heat_bias_pred}) and is the same model fitted in the case of unbiased sensor data (Section 6.2). The second model accounts for bias in the measurements by incorporating in the physics-informed prior a bias function as follows

\begin{equation*} 
\begin{split}
  y_u & = u(t,x) + \mathrm{Bias}(t,x) + \varepsilon_u, \text{ where } \mathrm{Bias}(t,x) \sim GP(0, K_{\mathrm{Bias}}((t,x),(t',x')))\\
  y_f &= f(t,x) + \varepsilon_f
\end{split}
\end{equation*}
and $K_{\mathrm{Bias}}((t,x),(t',x'))) = \sigma_B^2 \exp\left(-\frac{1}{2lB_t^2}(t-t')^2\right) exp\left(-\frac{1}{2lB_x^2}(x-x')^2\right).$ So we introduce to the model three additional hyperparameters ($\sigma_B, lB_t \text{ and } lB_x$).
	
In Figure \ref{fig:Post_Heat_Biased}, in the top row, we see the posteriors of the model that does not account for sensor bias. We observe that the physical parameter $\alpha$ is overestimated, and the posterior uncertainty ($90\%$ CI) does not cover the true value. The same holds for the heat noise parameter, $\sigma_u,$ and it captures the inability of the model to fit the observed data well, while for the unbiased forcing data, $f(t,x),$ the model estimates the noise parameter, $\sigma_f$ well with reasonable quantification of the uncertainty. In the second row of plots in Figure \ref{fig:Post_Heat_Biased}, we observe that the model that accounts for bias ($U(x,t)+\text{Bias}$) produces more reasonable quantification of uncertainty for $\alpha,$ and also the posterior density is concentrated very close to the true value.  The noise parameter, $\sigma_u,$ is underestimated. However, the true value is within the $90\%$ credible interval and also the posterior of the forcing noise parameter, $\sigma_f$ is almost identical to the model without bias.
	
In Figure \ref{fig:Heat_bias_pred}, we produce predictions for both models. In the first row ($u(t,x)$ model) we observe that when not accounting for bias, the model do not capture the true heat distribution shape, especially at the boundaries of the $x$ domain. By acknowledging in the model formulation that the data are biased  ($u(t,x)+\text{Bias}$ model) we see that the predictions capture the shape of the true heat distribution more accurately. However, this increases the uncertainty slightly in model predictions (shaded regions).
	
%---------------------------------------------
% Real case study

\section{Real data--WK models} \label{sec:real_data}
This case study is based in observations of blood flow and blood pressure from one individual that took part in a randomized controlled trial described in \cite{oyen2020effect}.  Our primary aim is to estimate the physical parameters vascular resistant ($R$)  and arterial compliance ($C$). 

\begin{figure}[h!]
	\centering
	\includegraphics[width=1\textwidth,height=0.25\textheight]{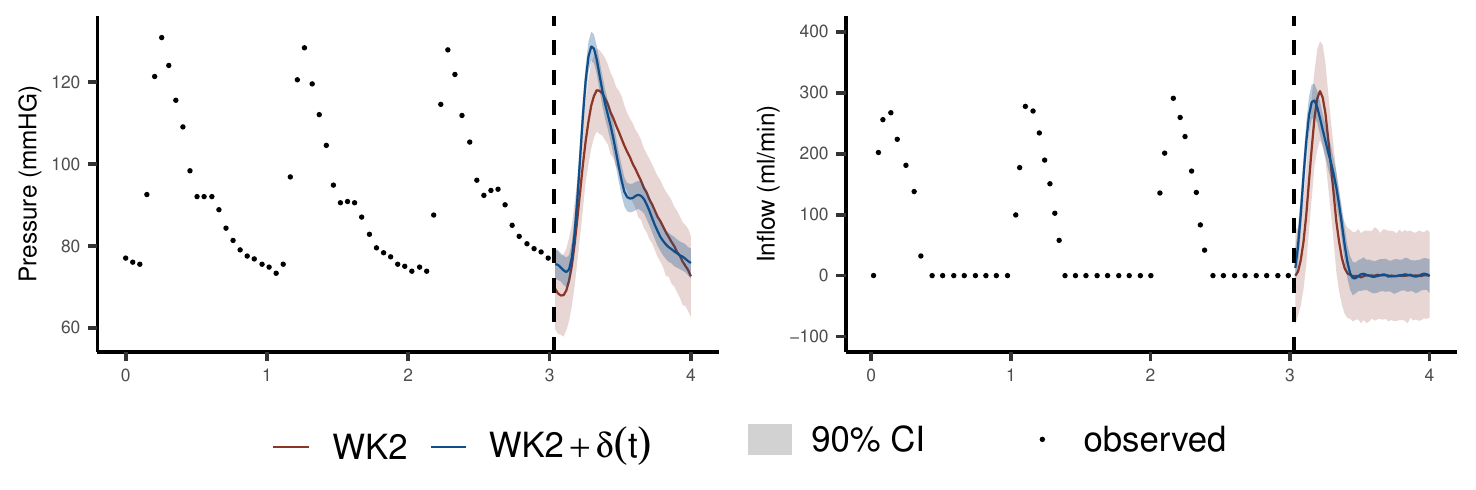}
	\vspace*{-4mm}	
	\caption{Predictions of blood pressure, $P(t)$ (left) and blood inflow, $Q(t)$ (right) for both models. The points represent the observed data, the solid lines are the mean predictions and the shaded regions are the $90\%$ credible intervals.}
	\label{fig:Pred_Real}
\end{figure}

The observations available are brachial blood pressure measured with Finometer PRO (Finapres Medical Systems, Enschede, Netherlands) on the right arm (see Figure \ref{fig:Pred_Real}, left) and blood inflow using Doppler flow (see Figure \ref{fig:Pred_Real}, right). We use three cycles for both pressure and flow. 

All analyses in this Section are based on is the WK2 model \eqref{eq:WK2} with physics-informed periodic kernel prior as described in Section \ref{sec:WK_synth}. 
The priors for the physical model parameters ($R$ and $C$) and the kernel hyperparameters are as in Section \ref{sec:WK_synth} with one exception, the observation noise prior $\sigma_u^2$. We know that the aortic valve is closed during diastole and the inflow is zero $Q(t)=0$.  In Figure \ref{fig:Pred_Real}, we find that the blood inflow is zero, $Q(t)=0$ for $\approx 2/3$ of each cardiac cycle. We introduce this knowledge into the model by setting the inflow noise, $\varepsilon_Q,$ to be 0 during the diastole;

\begin{equation*}
    \sigma_Q=
    \begin{cases}
    s_Q, \text { if } t=t_{\text{sys}}\\
    0, \text { if } t=t_{\text{dia}}
    \end{cases},
\end{equation*}
where $t_{\text{dia}}$ is for measurements during diastole and $t_{\text{sys}}$ is during systole.

We fit two models to these observations, the full Bayesian model (referred to as $\text{WK2}$) and the models  accounting for model discrepancy  (referred to as $\text{WK2}+ \delta$), as described and specified in Sections \ref{sec:Unbiased_WK2} and \ref{sec:Biased_model}. 

\begin{figure}[h!]
	\centering
	\includegraphics[width=1\textwidth,height=0.25\textheight]{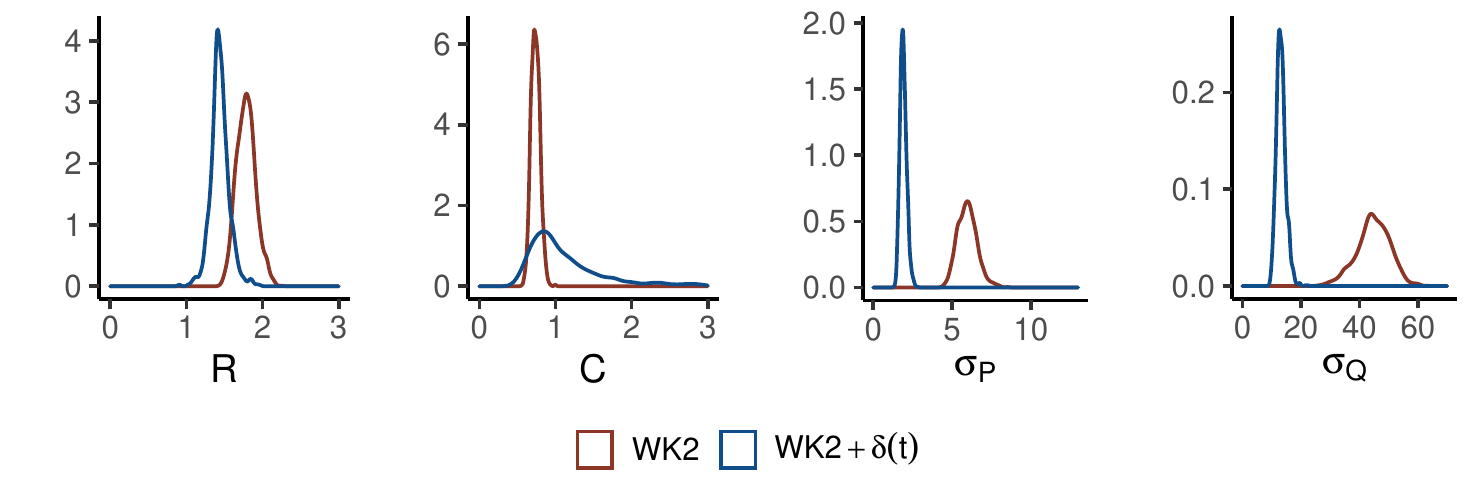}
	\vspace*{-4mm}	
	\caption{Posterior distributions for physical parameters ($R,C$) and noise parameters ($\sigma_P, \sigma_Q$) for the two models (WK2, $\text{WK2}+\delta(t)$).}
	\label{fig:Post_Real}
\end{figure}
The posterior distributions of the physical parameters ($R,C$) and noise parameters ($\sigma_P,\sigma_Q$) are found in Figure \ref{fig:Post_Real}, and blood inflow and pressure prediction with 90\% posterior prediction intervals are given in Figure \ref{fig:Pred_Real}.
The most striking differences are that  the noise parameters $\sigma_Q$ and $\sigma_P$  for the model without discrepancy ($\text{WK2}$) are much larger than for the model with discrepancy $\text{WK2}+\delta(t)$. The inflow noise standard deviation, $\sigma_Q$ for the WK2 model suggests that the observed inflow can be up to 60\% noise, which is not realistic in Figure \ref{fig:Pred_Real}, right. 
Further, the vascular resistant parameter $R$ is smaller for the model, including discrepancy.  For the arterial compliance parameter $C,$ the $\text{WK2}+\delta$ model gives larger uncertainty, and larger posterior mean then the $\text{WK2}$ model.  

In Figure \ref{fig:Pred_Real}, the predictions, as defined in Sections \ref{sec:FullBayes} and \ref{sec:BCPI_main}, for both pressure and inflow for the two models are plotted. We observe that the WK2 model doesn't reproduce the blood pressure waveform, and the prediction uncertainty is large, especially for the blood inflow. 
However, by accounting for model discrepancy ($\text{WK}+\delta$), the missing physics is learned from data, resulting in model predictions with reduced uncertainty in both pressure and inflow.

When comparing the result from fitting the WK2 and $\text{WK2}+\delta(t)$ models to the real data, we see the same pattern as in Section \ref{sec:Biased_model}. There synthetic data from WK3 models were fitted to WK2 models with and without discrepancy. We, therefore, find it reasonable to suspect that using a WK2 model without accounting for model discrepancy gives us too large $R$, overconfidence for $C$ and too large observation noise. 
%------------------------------------------------------------------

\section{Baseline comparison}\label{sec:comparison}

In this Section, the Bayesian calibration with physics-informed priors is compared with the two methods it is based on--1) the Bayesian calibration framework proposed by \citet{kennedy2001bayesian} (KOH) and 2) the physics-informed Gaussian process priors \citep{raissi2017machine}. We start by describing the main ingredients of the KOH modelling framework.

\noindent KOH modelled the noisy observed data $\mathbf{y}$ as follows
\begin{equation}
    y(\mathbf{x})=\eta(\mathbf{x},\boldsymbol \phi)+\delta(\mathbf{x})+\varepsilon,
\end{equation}
where $\eta$ is the physical model, $x$ is the observed inputs, $\boldsymbol{\phi}$ is the vector of physical parameters, $\delta$ is the model discrepancy and $\varepsilon$ is the noise term. 
A GP prior is also assumed on the model discrepancy $\delta(\mathbf{x})\sim GP(0,K_{\delta}(\mathbf{x},\mathbf{x}')).$ 
As mentioned in Section \ref{sec:KOH_intro}, KOH replace the physical model $\eta(\mathbf{x},\boldsymbol \phi)$ with an emulator trained on data obtained by the (physical model) numerical simulator on a $[\mathbf{x},\boldsymbol{\phi}]-$ space design. 
The emulator is another GP model trained on $N$ data points obtained by the simulator. 
Therefore, the final KOH model utilize two sources of information, $n$ observed data and $N$ simulator data. 
Hence the computational cost is $\mathcal{O}((N+n)^3),$ where typically $N\gg n.$ 

We consider a simulation study similar to Section \ref{sec:Biased_model}. 
More specifically, we simulate data from the WK3 model and use the WK2 as a modelling choice. 
Since the physical model consists of two functional outputs (blood pressure, $P(t)$ and blood flow $Q(t)$) we should use a multi-output GP emulator. However, this is not feasible due to the computational cost of the KOH model. 
Therefore, we assume blood flow is a known input to the model. %,  and we fix its values to the true ones. 
For the other two models, blood inflow is modelled as an output, observed with noise, as in Section \ref{sec:Biased_model}. 
Further, for the KOH approach, the number of pressure observations is reduced compared to Section \ref{sec:Biased_model} due to computational cost.

\begin{table}[ht] 
\centering
\resizebox{\columnwidth}{!}{%
\begin{tabular}{llllllll}
  \hline
 & R & C & $\sigma_P$ & $\sigma_Q$ & \# par &RMSE & runtime (sec)\\ 
  \hline
  True & 1.05 & 1.10 & 4 & 10 &  &\\ 
  \hline
  \hline
  PI opt & 0.93 (NA,NA) & 0.96 (NA,NA) & 7.89 (NA,NA) & 8.35 (NA,NA) & 6 & 6.85 & 2\\ 
  KOH & 1.15 (0.69,1.65) & 1.04 (0.94,1.13) & 3.94 (3.17,4.88) & NA (NA,NA) & 11 & 1.91 & 483\\ 
  \textbf{BCPI} & \textbf{1.06} (0.8,1.46) & \textbf{1.08} (0.73,1.74) & 4.10(3.21,5.19) & 8.68 (7.09,10.69) & 8 & \textbf{1.61} & 23\\ 
   \hline
\end{tabular}
}
\caption{Baseline comparison results. PI opt is the physics-informed prior model proposed by \cite{raissi2017machine}. KOH is the Bayesian calibration approach proposed by \cite{kennedy2001bayesian}. BCPI is the proposed method, Bayesian calibration with physics-informed priors. The PI opt model provides only point estimates of the model parameters, while both the KOH and proposed method (BCPI) provide the posterior distribution of the parameters. The point estimates for KOH and proposed (BCPI) are the posterior means. The \# par is the number of parameters for each model.}\label{table:comparison}
\end{table}

In order to build an emulator for the KOH model, we use as response the simulated blood pressure data, $P^{\text{WK2}}$ from the WK2 model and as inputs to the emulator model the blood inflow $Q,$ time $t_P$ and the physical parameters $R$ and $C.$ To simulate blood pressure data, we run the model on 12 different physical parameter values obtained from an experimental design on the space $R\times C = [0.5,3]^2.$ 
Note that the experimental design values can greatly influence the results of the KOH approach. 
This is a strength of the BCPI method since it uses only observed data and therefore does not need any simulator data. 
Another strength is that it does not need any initial or boundary conditions in order to build a numerical simulator for the physical model, which in practice it might be hard to know.

The results are summarized in Table \ref{table:comparison}. 
The estimated values of the physical parameters $R, C$ and the noise parameters $\sigma_P, \sigma_Q$ as well as the prediction root mean square error (RMSE) and the runtime are presented. 
The PI opt model is the approach proposed by \cite{raissi2017machine}, where the parameters of the physics-informed GP prior are optimized. 
Therefore the uncertainty of the physical parameters is not included, and the model also does not account for model discrepancy. 
We observe that the model produces biased estimates for both $R$ and $C$ and also overestimates the pressure noise parameter $\sigma_P,$  which is similar to the Bayesian alternative of the model  (see Figures \ref{fig:Post_all_kernels_Biased} and \ref{fig:range_R1}). 
The KOH model produces more reliable estimates of the physical parameters, where the posterior distribution covers the true values. 
However, this comes with a computational cost. The PI opt model takes 2 seconds to run, while the KOH model takes 483 seconds. 
Observe also that the KOH model, in this case, does not account for the uncertainty in the inflow data. 
The proposed approach, Bayesian calibration with physics-informed priors (BCPI) produces more reliable parameter estimates. 
It takes only 23 seconds to run, which is a considerable reduction compared to the KOH model, while it also models uncertainty in both outputs ($P(t)$ and $Q(t)$). 
Note that the KOH model has 11 unknown parameters while the proposed approach has 8. 
The extra two parameters compared to the PI opt model are the parameters in the discrepancy process. 
To make runtimes comparable, we use HMC sampling implemented in STAN, sampling three chains of 1000 samples in parallel for both the KOH and the proposed method (BCPI) models. 
Finally, we see that the PI opt model, which does not account for model discrepancy, produces unreliable predictions having an RMSE of 6.85. The KOH has a much lower RMSE of 1.91, while the proposed method (BCPI) has the smallest RMSE (1.61) of all three methods. 

There are many alternatives or improvements to the KOH approach that we do not include in this comparison. 
For example, a commonly used approach in practice is the modularized KOH approach \citep{bayarri2009modularization} or a recently developed method that uses deep GPs \citep{marmin2022deep}, which might improve the standard KOH approach through more complex modelling structures. 
An important difference between BCPI and the methods mentioned above and in Section \ref{sec:KOH_intro} is that it does not need any simulator data to fit the model. However, the ideas mentioned above or other ideas on how to deal with big data can be applied to our approach. 
For example, in Section \ref{sec:bigdata} we develop two approximations for big data based on two popular GP methods for big data. 

\section{Approximations for big data} \label{sec:bigdata}

In this Section, we develop two approximations for the physics-informed prior models described in Section \ref{sec:method}. 
These approximations are based on two of the most influential GP models for big data, the Fully Independent Training Conditional (FITC) \citep{snelson2005sparse} and the Variational Free Energy (VFE) \citep{titsias2009variational}.

Compared to the standard flexible GP models, the physics-informed GP priors are quite informative since they are constructed in a way that they satisfy the differential equation. More specifically, the models incorporate information about the physical process in the covariance (and the mean) function. 
Therefore, our assumption is that we do not need a large number of data to reliably infer the latent functions and the physical model parameters. 

A similar assumption is made in two of the most popular GP approaches for big data.
The FITC and VFE approximations assume that most of the observed data are redundant and reduce the effective number of input data from $N$ to $m,$ where $m \ll N$. 
The $m$ data points are called inducing points (or pseudo inputs), and both methods reduce the computational cost from  $\mathcal{O}(N^3)$ to $\mathcal{O}(N\cdot m^2).$

In Section \ref{sec:FITC_VFE}, we derive the physics-informed FITC and VFE approximations and the predictive equations. 
Sections \ref{sec:bigdata_exper_noDelta} and \ref{sec:bigdata_exper_withDelta}, consider two experiments with the Windkessel models, with and without model discrepancy, respectively.

\subsection{Physics-informed FITC and VFE approximations} \label{sec:FITC_VFE}

In the regression setting, we model the latent function $g(\cdot)$ using a zero mean GP prior, $\mathbf{g}\sim GP(0,K)$ for which we have noisy observed outputs $y_i,i=1,\ldots,N,$ at the input locations $\mathbf{x}_i,i=1,\ldots,N.$ We assume Gaussian i.i.d. noise $\varepsilon\sim N(0,\sigma_n^2I),$ and we have that  $p(\mathbf{g}) = \mathcal{N}(0, K_{\mathbf{g}\mathbf{g}})$ and $p(\mathbf{y}\mid \mathbf{g}) = \mathcal{N}(\mathbf{g}, \sigma_n^2I).$ 
The FITC and VFE approximations introduce a set of $m$ inducing variables $\mathbf{w}=(w_1,\ldots,w_m)$ at the corresponding inputs $\mathbf{Z} = (\mathbf{z}_1,\ldots, \mathbf{z}_m),$ where $w_i=g(\mathbf{z}_i).$ 
As in the standard GP model, inference is based on the log marginal likelihood, which is given by the following expression for both models \citep{bauer2016understanding}
\begin{equation}
    L = \log\mathcal{N}(0, Q_{\mathbf{g}\mathbf{g}}+\Lambda) -\frac{1}{2\sigma_n^2} \text{tr}(T),
\end{equation}
where $Q_{\mathbf{g}\mathbf{g}} = K_{\mathbf{g}\mathbf{w}}K_{\mathbf{w}\mathbf{w}}K_{\mathbf{w}\mathbf{g}}$ is a low-rank matrix, which reduces size of the matrix inversion from $N$ to $m$. The terms $\Lambda$ and $T$ differ between the two models and are given as follows
\begin{align}
    \Lambda_\text{FITC} &= \textrm{diag}(K_{\mathbf{g}\mathbf{g}}- Q_{\mathbf{g}\mathbf{g}}) + \sigma_n^2I &  \quad T_\text{FITC}&=0 \label{eq:FITC} \\
    \Lambda_\text{VFE} &=  \sigma_n^2I & \quad  T_\text{VFE}&= K_{\mathbf{g}\mathbf{g}}- Q_{\mathbf{g}\mathbf{g}}. \label{eq:VFE}
\end{align}
The prediction equations at new points $\mathbf{X^*},$ $g(\mathbf{X}^*) = \mathbf{g}^*$  are given for both FITC and VFE by the following expression
\begin{equation}
    p(\mathbf{g}^*) = \mathcal{N}(\mu^*\,,\, \Sigma^*)
\end{equation}
\begin{equation}
\begin{aligned}
\mu^* &= \mu(X^*) + K_{*\mathbf{w}}(K_{\mathbf{w}\mathbf{w}} + K_{\mathbf{w}\mathbf{g}} \Lambda^{-1} K_{\mathbf{g}\mathbf{w}})^{-1} \Lambda^{-1}(\mathbf{y} - \boldsymbol{\mu}) \\
\Sigma^* &= K_{**} - K_{*\mathbf{w}}K_{\mathbf{w}\mathbf{w}}K_{\mathbf{w}*} + K_{*\mathbf{w}}(K_{\mathbf{w}\mathbf{w}} + K_{\mathbf{w}\mathbf{g}}\Lambda^{-1} K_{\mathbf{g}\mathbf{w}})^{-1}K_{\mathbf{w}*},
\end{aligned}
\end{equation}
where $\Lambda$ as given as in equations \eqref{eq:FITC} and \eqref{eq:VFE} for FITC and VFE respectively. 

The models in Section \ref{sec:method} are multi-output Gaussian process models. 
Sparse approximations for multi-output (or multi-task) GPs have already been introduced in the literature \citep{alvarez2008sparse,alvarez2010efficient}, with kernels based on convolution processes. The main difference is that our kernels are based on the differential equation that the multi-output process describes. 

To derive approximations for the physics-informed GP models, we keep the notation similar to the standard GP model described above but with some differences. 
Recall that the models are built for the differential equations $\mathcal{L}_{x}^{\boldsymbol\phi}u(x) = f(x).$ 
Therefore, the vector of latent variables $\mathbf{g}$ now represents the two vectors $\mathbf{u}$ and $\mathbf{f}$ as  $\mathbf{g} = (\mathbf{u},\mathbf{f}).$ 
Similarly, we consider $m_u$ inducing variables $\mathbf{w}_u$ for the function $u(\cdot)$ and $m_f$ inducing variables $\mathbf{w}_f$  for the function $f(\cdot),$ $\mathbf{w}=(\mathbf{w}_u,\mathbf{w}_f)$ at the input locations $\mathbf{Z} = (\mathbf{Z}_u, \mathbf{Z}_f),$ where $w_{ui} = u(Z_{ui})$ and $w_{fj} = f(Z_{fj}).$ The covariance function $K$ of the FITC and VFE approximations is now replaced by the physics-informed covariance function of Section \ref{sec:method}  
\begin{equation}\label{eq:PI_cov}
   K^{\text{PI}} =
   \begin{bmatrix}
   K_{uu}(\mathbf{X}_u, \mathbf{X}_u)  & K_{uf}(\mathbf{X}_u, \mathbf{X}_f)\\
   K_{fu}(\mathbf{X}_f, \mathbf{X}_u) & K_{ff}(\mathbf{X}_f, \mathbf{X}_f)
\end{bmatrix}, 
\end{equation}
where the kernel hyperparameters are dropped for notational convenience. The marginal log likelihood is given by the following expression
\begin{equation}
    L^{\text{PI}} = \log\mathcal{N}(\boldsymbol{\mu}^\text{PI}, Q^{\text{PI}}_{\mathbf{g}\mathbf{g}}+\Lambda^{\text{PI}}) -\text{tr}(S^{-1}T^{\text{PI}}),
\end{equation}
where $Q^{\text{PI}}_{\mathbf{g}\mathbf{g}} = K^{\text{PI}}_{\mathbf{g}\mathbf{w}}K^{\text{PI}}_{\mathbf{w}\mathbf{w}}K^{\text{PI}}_{\mathbf{w}\mathbf{g}}$ is a low-rank matrix, which reduces size of the matrix inversion from $N_u+N_f$ to $m_u+m_f,$ and
\begin{equation}
   K^{\text{PI}}_{\mathbf{w}\mathbf{w}} =
   \begin{bmatrix}
   K_{\mathbf{u}\mathbf{u}}(\mathbf{Z}_u, \mathbf{Z}_u)  & K_{\mathbf{u}\mathbf{f}}(\mathbf{Z}_u, \mathbf{Z}_f)\\
   K_{\mathbf{f}\mathbf{u}}(\mathbf{Z}_f, \mathbf{Z}_u) & K_{\mathbf{f}\mathbf{f}}(\mathbf{Z}_f, \mathbf{Z}_f)
\end{bmatrix}
\text{ and }
   K^{\text{PI}}_{\mathbf{g}\mathbf{w}} =
   \begin{bmatrix}
   K_{\mathbf{u}\mathbf{u}}(\mathbf{X}_u, \mathbf{Z}_u)  & K_{\mathbf{u}\mathbf{f}}(\mathbf{X}_u, \mathbf{Z}_f)\\
   K_{\mathbf{f}\mathbf{u}}(\mathbf{X}_f, \mathbf{Z}_u) & K_{\mathbf{f}\mathbf{f}}(\mathbf{X}_f, \mathbf{Z}_f)
\end{bmatrix}.
\end{equation}
The terms $\Lambda^{\text{PI}}$ and $T^{\text{PI}}$ for the two models and are given as follows
\begin{align}
    \Lambda^{\text{PI}}_\text{FITC} &= \textrm{diag}(K^{\text{PI}}_{\mathbf{g}\mathbf{g}}- Q^{\text{PI}}_{\mathbf{g}\mathbf{g}}) + S &  \quad T^{\text{PI}}_\text{FITC} &=0 \label{eq:FITC_PI} \\
    \Lambda^{\text{PI}}_\text{VFE}  &=  S & \quad  T^{\text{PI}}_\text{VFE}&= K^{\text{PI}}_{\mathbf{g}\mathbf{g}}- Q^{\text{PI}}_{\mathbf{g}\mathbf{g}}, \label{eq:VFE_PI}
\end{align}
where 
$
\mathbf{S} =
\begin{bmatrix}
\sigma_u^2 I_u & 0\\
0 & \sigma_f^2 I_f
\end{bmatrix}.
$
Note that if we account for model discrepancy or biased data, we replace the covariance matrix $K_{\mathbf{u}\mathbf{u}}$ of equation \eqref{eq:PI_cov} with $K_{\mathbf{u}\mathbf{u}}+ K_\delta$ and $K_{\mathbf{u}\mathbf{u}}+ K_\text{Bias}$ respectively. The physics-informed FITC and VFE approximations reduce the computational cost from $\mathcal{O}((N_u+N_f)^3)$ to $\mathcal{O}((N_u+N_f)\cdot(m_u+m_f)^2),$ where $m_u \ll N_u$ and $m_f \ll N_f,$ and $N_u,N_f$ are the number of data for the functions $u$ and $f$ respectively. 

To make predictions at new points $X^*_u,$ $u(X^*_u) = \mathbf{u}^*,$ the predictive distribution is multivariate Gaussian and more specifically 
\begin{equation}
    p(\mathbf{u}^*) = \mathcal{N}(\mu^*\,,\, \Sigma^*)
\end{equation}
\begin{equation}
\begin{aligned}
\mu^* &= \mu(X^*_u) + \mathbf{V}_{w_u}^*{^T}(K_{\mathbf{w}\mathbf{w}} + K_{\mathbf{w}\mathbf{g}} \Lambda^{-1} K_{\mathbf{g}\mathbf{w}})^{-1} \Lambda^{-1}(\mathbf{y} - \boldsymbol{\mu}) \\
\Sigma^* &= K_{**} - \mathbf{V}_{w_u}^*{^T}K_{\mathbf{w}\mathbf{w}}\mathbf{V}_{w_u}^* + \mathbf{V}_{w_u}^*{^T}(K_{\mathbf{w}\mathbf{w}} + K_{\mathbf{w}\mathbf{g}}\Lambda^{-1} K_{\mathbf{g}\mathbf{w}})^{-1}\mathbf{V}_{w_u}^*,
\end{aligned}
\end{equation}
where $\mathbf{V}_{w_u}^*{^T} = \begin{bmatrix}  K_{uu}(\mathbf{X}^*_u,\mathbf{Z}_u) &  K_{uf}(\mathbf{X}^*_u,\mathbf{Z}_f) \end{bmatrix}.$ 
Including model discrepancy in the formulation gives
$\mathbf{V}_{w_u}^*{^T} = \begin{bmatrix}  K_{uu}(\mathbf{X}^*_u,\mathbf{Z}_u)+K_{\delta}(\mathbf{X}^*_u,\mathbf{Z}_u) &  K_{gf}(\mathbf{X}^*_u,\mathbf{Z}_f) \end{bmatrix}$. 
Note that the superscript PI is dropped for notational convenience.
Similarly, if we want to make predictions at new points $X^*_f,$ $f(X^*_f) = \mathbf{f}^*,$ the predictive distribution is multivariate Gaussian and more specifically 
\begin{equation}
    p(\mathbf{f}^*) = \mathcal{N}(\mu^*\,,\, \Sigma^*)
\end{equation}
\begin{equation}
\begin{aligned}
\mu^* &= \mu(X^*_f) + \mathbf{V}_{w_f}^*{^T}(K_{\mathbf{w}\mathbf{w}} + K_{\mathbf{w}\mathbf{g}} \Lambda^{-1} K_{\mathbf{g}\mathbf{w}})^{-1} \Lambda^{-1}(\mathbf{y} - \boldsymbol{\mu}) \\
\Sigma^* &= K_{**} - \mathbf{V}_{w_f}^*{^T}K_{\mathbf{w}\mathbf{w}}\mathbf{V}_f^* + \mathbf{V}_{w_f}^*{^T}(K_{\mathbf{w}\mathbf{w}} + K_{\mathbf{w}\mathbf{g}}\Lambda^{-1} K_{\mathbf{g}\mathbf{w}})^{-1}\mathbf{V}_{w_f}^*,
\end{aligned}
\end{equation}
where $\mathbf{V}_{w_f}^*{^T} = \begin{bmatrix}  K_{fu}(\mathbf{X}^*_f,\mathbf{Z}_u) &  K_{ff}(\mathbf{X}^*_f,\mathbf{Z}_f) \end{bmatrix}.$

\subsection{Experiments: Full Bayes model without discrepancy}\label{sec:bigdata_exper_noDelta}
We consider a simulation study similar to Section \ref{sec:Unbiased_WK2}, where the physical parameters are $R=1,$ $C=1.1,$ and the noise parameters are $\sigma_P=4$ and $\sigma_Q=10.$ 
We assume that we have 100 inflow observations, $n_Q=100$ and 90 blood pressure observations $n_P=90.$ 
This amount of data could be handled by the methods described in Section \ref{sec:method}, though here, it is considered for illustration purposes. 
For a given inflow, $Q(t),$ we simulate pressure data from the deterministic WK2 model, $P^{\text{WK2}},$ and we add to both i.i.d. zero mean Gaussian noise, as in Section \ref{sec:Unbiased_WK2}. We fit the physics-informed prior for the WK2 model as in Section \ref{sec:Unbiased_WK2} using the FITC and VFE approximations derived in Section \ref{sec:FITC_VFE}. Eight inducing points for the blood pressure, $m_P=8$ and ten inducing points for the blood inflow, $m_Q=10$ are used for both the FITC and VFE approximations.
\begin{figure}[h!]
	\centering
	\includegraphics[scale=0.7]{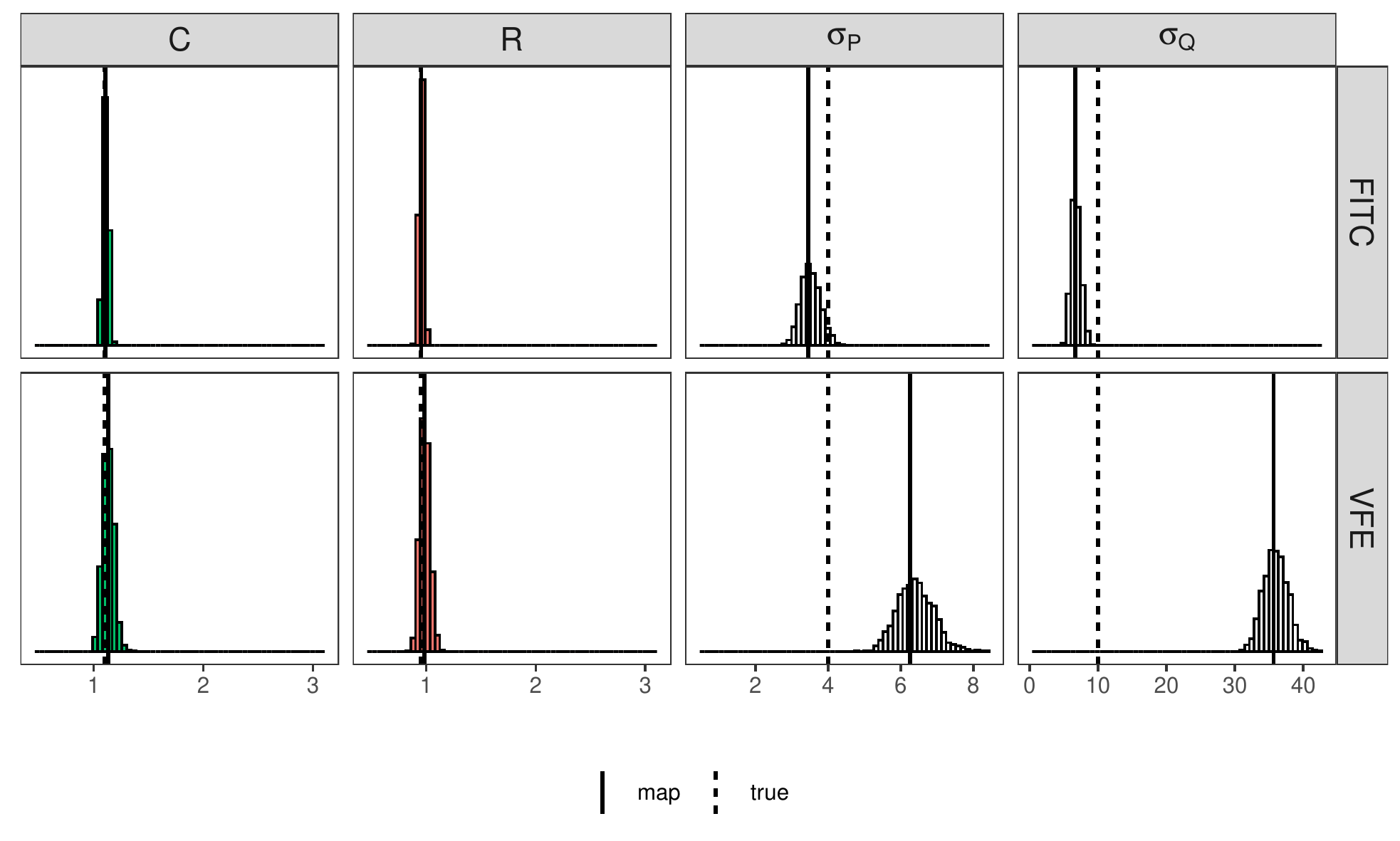}
	\vspace*{-4mm}	
	\caption{Physics-informed FITC and VFE models without discrepancy; Posterior distributions.}
	\label{fig:post_appr_nodelta}
\end{figure}

To fit the models, a modular approach is used. First, we optimize the marginal log-likelihood with respect to kernel hyperparameters, and the inducing point locations. 
Then we fix the inducing locations at the estimated values and sample the physic-informed parameters using HMC sampling as in Section \ref{sec:Unbiased_WK2}. 
The posterior distributions of the physical and noise parameters, along with point estimates from the first step, are presented in Figure \ref{fig:post_appr_nodelta} for the FITC and VFE approximations. 
Both models estimate the physical parameters accurately with relatively small uncertainty and is similar to Figure \ref{fig:Post_all_kernels_Unbiased}. 
Observe that the MAP (maximum a posteriori) estimate is quite accurate. 
However, the FITC model underestimates the noise while the VFE overestimates the noise parameters, which are known characteristics of the two approximations \citep{bauer2016understanding}.
\begin{figure}[h!]
	\centering
	\includegraphics[scale=0.75]{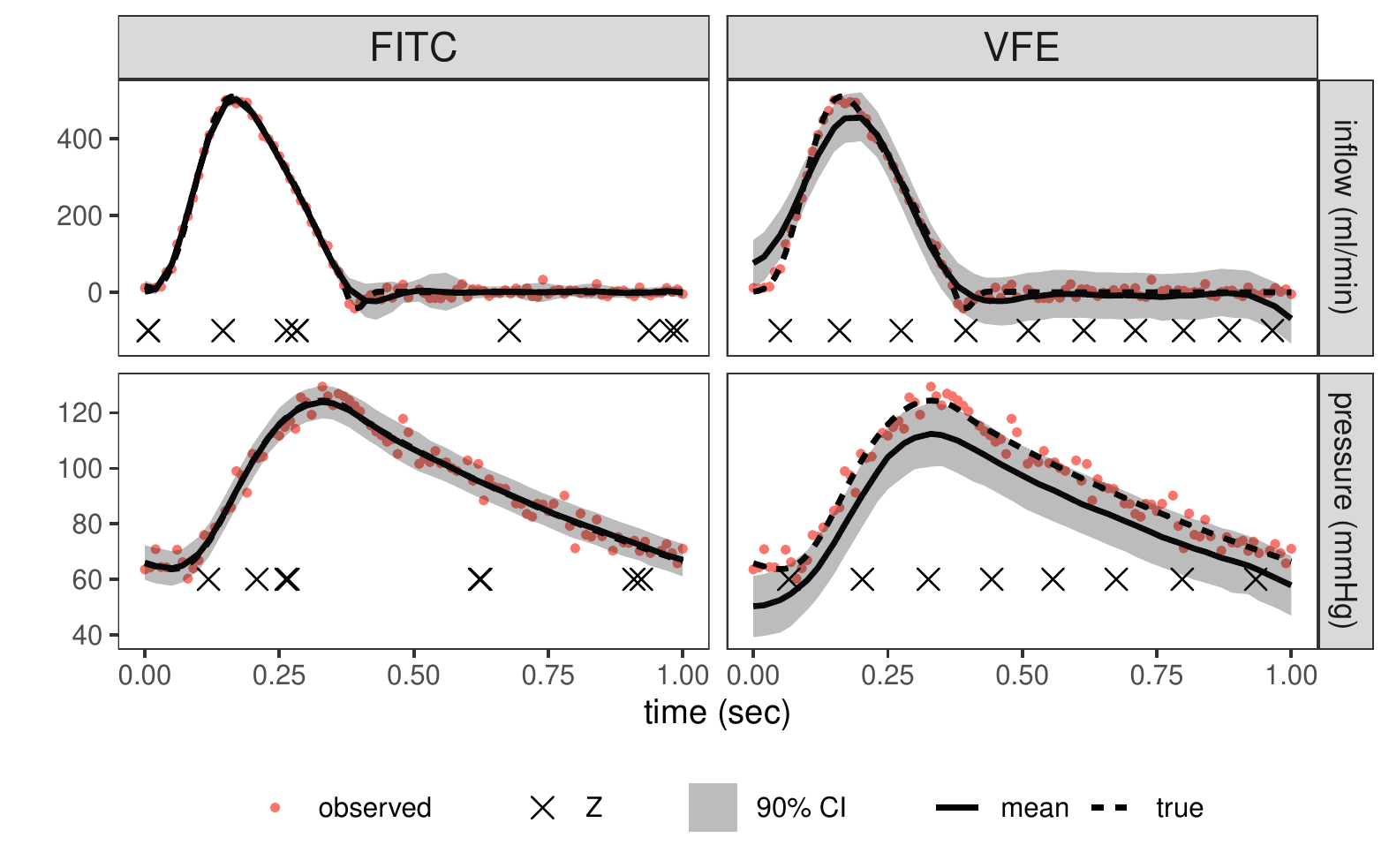}
	\vspace*{-4mm}	
	\caption{Physics-informed FITC and VFE models without discrepancy; Predictions.}
	\label{fig:pred_appr_nodelta}
\end{figure}
\begin{figure}[h!]
	\centering
	\includegraphics[scale=0.75]{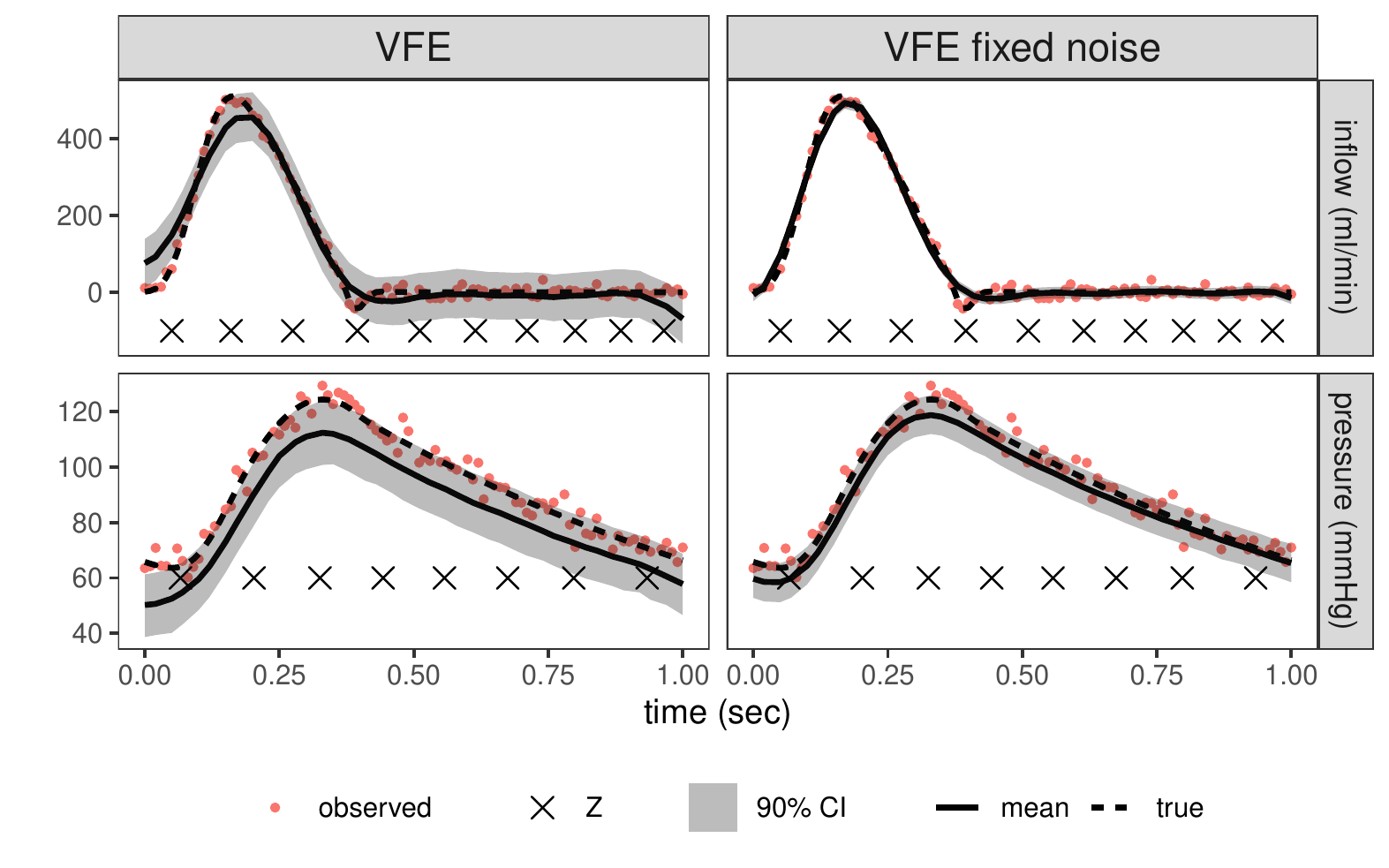}
	\vspace*{-4mm}	
	\caption{Physics-informed VFE without discrepancy; Predictions with fixed noise.}
	\label{fig:pred_appr_nodelta_plugin}
\end{figure}

We use the posterior distributions of the parameters and the fixed inducing locations to produce predictions.  
The prediction means, along with the $90\%$ credible intervals and the inducing input locations for both functions, are presented in Figure \ref{fig:pred_appr_nodelta}. 
The FITC model produces accurate predictions for both functions, where the heteroscedastic nature of the predictions can be a desired property. 
The VFE model underfits both inflow and blood pressure with relatively large prediction uncertainty. 
This is also a known characteristic of the VFE approximation \citep{lazaro2009inter}, though this might be an optimization issue \citep{bauer2016understanding}. 
As a remedy to this problem, we use a sample of the observed blood pressure and inflow data and we fit independent standard GP models to obtain point estimates of the noise parameters $\sigma_u$ and $\sigma_f.$ Then, we predict with the VFE model by fixing the noise parameter values to the estimated ones. 
In Figure \ref{fig:pred_appr_nodelta_plugin}, right, we see that the prediction accuracy of the VFE model has significantly improved compared to using the overestimated noise parameter values (same Figure left).

\subsection{Experiments: Accounting for model discrepancy}\label{sec:bigdata_exper_withDelta}
We consider a similar experimental setup to Section \ref{sec:bigdata_exper_noDelta}, but now we simulate from a more complex model than our modelling choice. More specifically, as in Section \ref{sec:Biased_model}, for a given inflow $Q(t),$ we simulate data from the deterministic WK3 model, $P(t)=P_{\text{WK3}}(Q(t), R_1=0.05, R_2=1, C=1.1),$ and we add i.i.d. Gaussian noise as in Section \ref{sec:bigdata_exper_noDelta}. Our modelling choice is the WK2 model with model discrepancy ($\text{WK2}+\delta(t)$) as in Section \ref{sec:Biased_model}, and we use 12 inducing points for the blood pressure $m_P=12$ and 10 inducing points for inflow $m_Q=10.$

\begin{figure}[h!]
	\centering
	\includegraphics[scale=0.7]{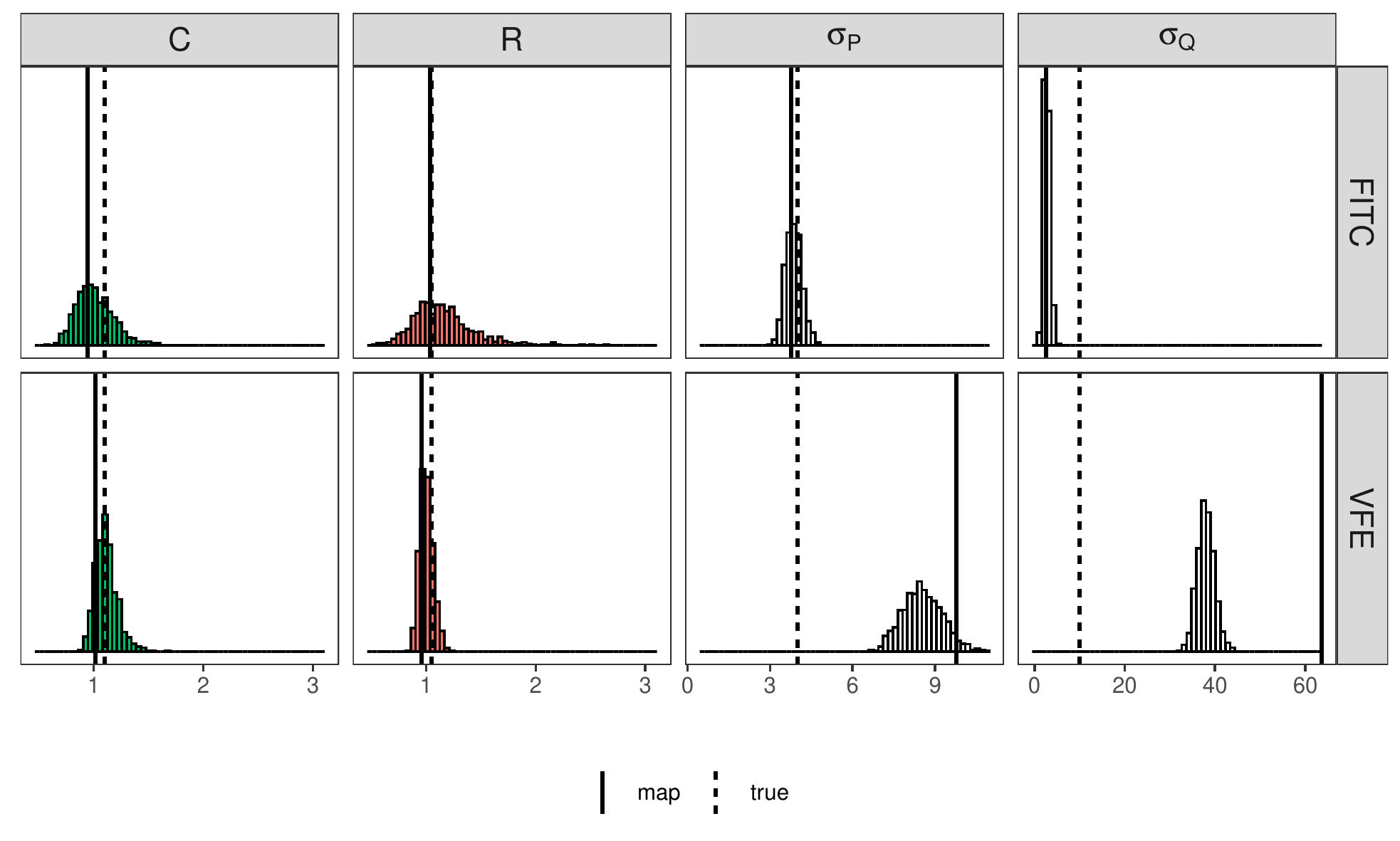}
	\vspace*{-4mm}	
	\caption{Physics-informed FITC and VFE models with discrepancy; Posterior distributions.}
	\label{fig:post_appr_delta}
\end{figure}

To fit the models, we use the same modular approach as in Section \ref{sec:bigdata_exper_noDelta}. 
The posterior distributions of the physical and noise parameters for both models, along with the point estimates of the optimization step, are presented in Figure \ref{fig:post_appr_delta}. 
The posterior distributions of physical parameters cover the true values for both models, where the posterior uncertainty for the VFE model is smaller. 
The MAP estimates of the physical parameters $R$ and $C$ are also quite close to the true value. Hence in cases where the data size is quite large, and MCMC is not feasible, MAP estimates might be a practical solution. The FITC model estimates the pressure noise parameter $\sigma_P$ accurately while it underestimates the inflow noise parameter, $\sigma_Q,$ again. As is Section \ref{sec:bigdata_exper_noDelta}, the VFE model overestimates the noise for blood pressure and inflow.
 
\begin{figure}[h!]
	\centering
	\includegraphics[scale=0.75]{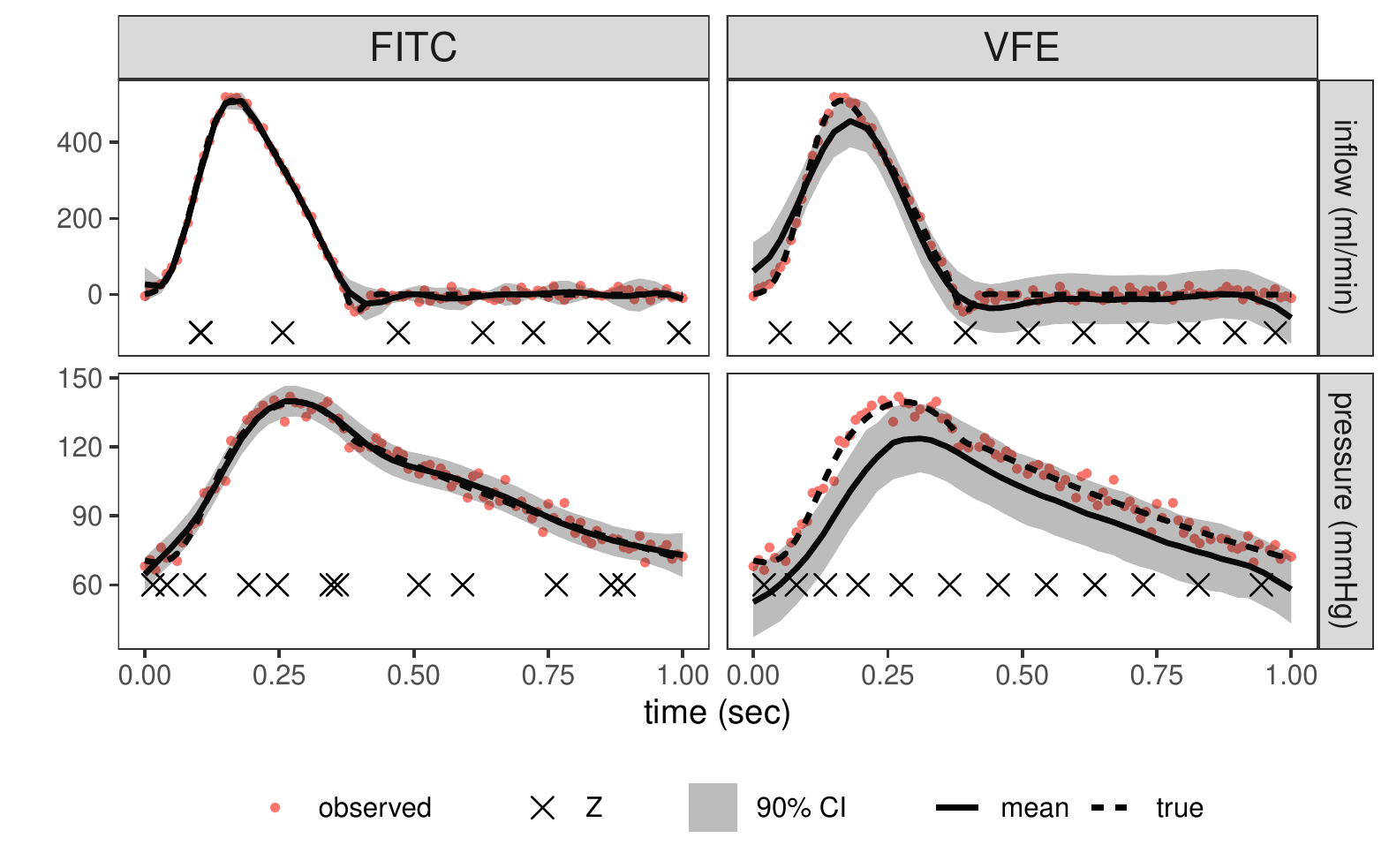}
	\vspace*{-4mm}	
	\caption{Physics-informed FITC and VFE models with discrepancy; Predictions.}
	\label{fig:pred_appr_delta}
\end{figure}

\begin{figure}[h!]
	\centering
	\includegraphics[scale=0.75]{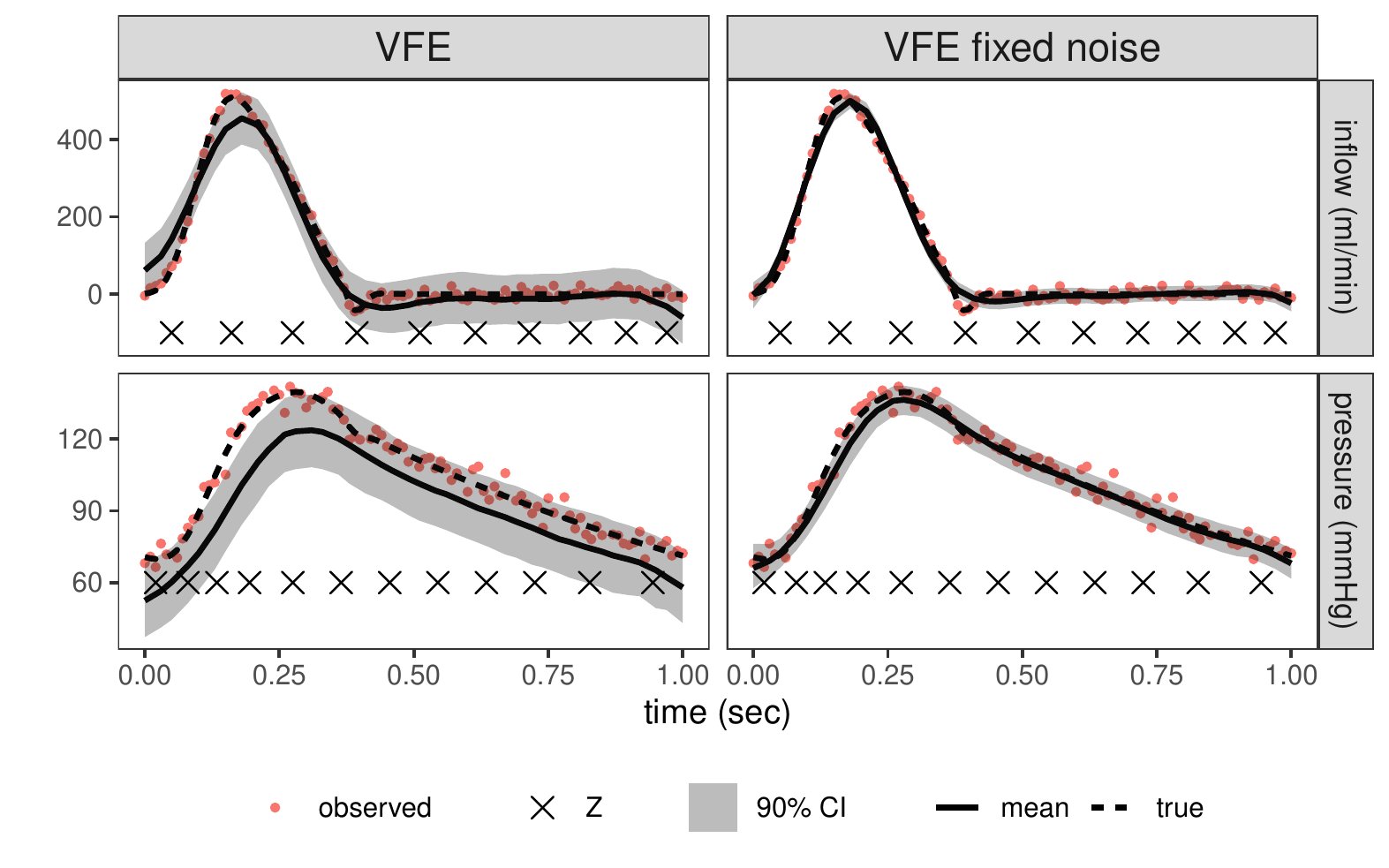}
	\vspace*{-4mm}	
	\caption{Physics-informed FITC and VFE models with discrepancy; Predictions with plug in point estimates for noise parameters.}
	\label{fig:pred_appr_delta_plugin}
\end{figure}

Further predictions are produced based on the posterior distributions of the parameters and the fixed inducing locations. 
The prediction means, along with the $90\%$ credible intervals and the inducing point locations for the FITC and VFE approximations, are presented in Figure \ref{fig:pred_appr_delta}. 
The FITC approximation produces accurate predictions for both blood pressure and inflow, while the VFE approximation does not fit the data well with large prediction uncertainty as in Section \ref{sec:bigdata_exper_noDelta}. 
To improve predictions of the VFE model, we use the same approach as in Section \ref{sec:bigdata_exper_noDelta}. 
First, independent standard GP models are fitted on a sample of blood pressure and inflow data and point estimates of $\sigma_P$ and $\sigma_Q$ are obtained. 
Predictions are obtained using the posteriors of the parameters and the fixed inducing locations, but now we fix the noise parameters to the point estimates. 
Comparing the right and left plot in Figure \ref{fig:pred_appr_delta_plugin}, we see that by fixing the noise parameters to more reasonable values, the VFE model can produce much more reliable predictions.

\section{Discussion and Conclusion} \label{sec:conc}
We have presented a Bayesian framework for calibration of computer models represented by differential equations of the following form, $\mathcal{L}_{x}^{\boldsymbol\phi}u(x) = f(x),$ using physics-informed priors. 
Compared to other Bayesian calibration frameworks, our approach is more exact in the sense that we do not use a model emulator, which is an approximation to the physical model trained on data obtained by simulations. 
Rather, we use a physics-informed prior, a probabilistic model that satisfies the differential equation. 
This also gives a computational advantage since we do not have to carry through inference from simulations of the differential equation which is often the main computational bottleneck. 
Instead, the model is evaluated on observed data only. 
We took a fully Bayesian approach using HMC sampling for learning the model parameters since our primary interest is learning the uncertainty of physical parameters. 

The computational cost of the proposed methods is $\mathcal{O}(N^3),$ where $N$ is the  total number of observations of functions $u$ and $f,$ and in applications of physical models the number of observed data is typically small. However, in cases where $N$ is large, the cubic cost is prohibitive. For this reason, we derived approximations for our method that reduce the computational cost to $\mathcal{O}(N\cdot m^2),$ where $m\ll N.$ We also found in experimental studies that the approximations produce accurate estimates of the physical parameters and predictions. 

Our approach can be generalized for systems of differential equations, based on \citet{sarkka2011linear}. When the physical model is described by non-linear differential equations, recent ideas on how to construct physics-informed priors for non-linear ODEs and PDEs can be used \citep{raissi2018numerical,chen2022apik}.

We demonstrated the flexibility of this approach using a time-dependent ODE, the arterial Windkessel model, and a space-time PDE, the heat equation, for both real and simulated data in cases of model discrepancy and biased sensor data. In a simulation study we demonstrated that by accounting for model discrepancy in a low fidelity model we could recover the true parameter values of a more complex model and produce more accurate predictions. In the case of biased sensor data, we showed that by accounting for this bias in the model formulation, we could recover the true value of the physical parameter (diffusivity constant) and produce more reliable model predictions. However, uncertainty is not reduced because the bias that the model learned should be removed in predictions compared to the model discrepancy case where we learn the missing physics and use this information in predictions. 

In applications, we might have to deal with both model discrepancy and biased data. 
In such cases, the model should account for both sources of uncertainty and can be written as $y(\mathbf{x})=\eta(\mathbf{x},\boldsymbol{\phi})+\delta(\mathbf{x})+\textrm{Bias}(\mathbf{x})+\varepsilon.$ 
If we assume flexible noninformative GP priors for both the discrepancy, $\delta \sim GP(0,K_\delta(\mathbf{x}, \mathbf{x}'))$ and $\textrm{Bias} \sim GP(0,K_{\textrm{Bias}}(\mathbf{x}, \mathbf{x}'))$ there will be identifiability issue between the two processes and, therefore it will be hard to separate $\delta$ from $\textrm{Bias}.$ 
In cases that we have prior information for $\delta$ or $\textrm{Bias},$ we might use informative priors to improve identifiability or a restrictive parametric form \citep{oliver2018calibration}. 
If there is no prior information available, and hence the two processes can not be separated, we can use a flexible GP to account for both $\delta$ and $\textrm{Bias},$ which results in the original KOH formulation, 
$y(\mathbf{x})=\eta(\mathbf{x},\boldsymbol{\phi})+\delta(\mathbf{x})+\varepsilon.$ 
Therefore, the term $\delta$ will absorb the effect of both the discrepancy and data bias, and there will be able to recover the physical parameter values. However, this model can not be used for predictions since the data bias can not be removed from the predictive equations.

A potential issue in Bayesian calibration is the identifiability between the discrepancy function and model parameters. 
The requirements for identifiability have been studied by \citet{arendt2012quantification}, showing that identifiability can be achieved under the mild assumption of a smooth discrepancy function. 
\citet{arendt2012improving} showed that using multiple functions that they depend on the same set of physical parameters can improve identifiability, or in other words using a multi-output GP model. Note that the models of the proposed method are by default multi-output GPs. 
Other ideas for enhancing identifiability in the KOH formulation include the introduction of shape constraints in the discrepancy function when prior information is available \citep{brynjarsdottir2014learning,riihimaki2010gaussian, wang2016estimating} or forcing the discrepancy function $\delta(\cdot)$ to be orthogonal to the emulator of the physical model \citep{plumlee2017bayesian}. Both ideas can be incorporated to our modelling framework.

\newpage

\section*{Appendix}
\appendix

\section{Prediction equations}
In general if $\mathbf{f}\sim GP(\mathbf{\mu(\mathbf{X})}, K(\mathbf{X,X'})),$ at new points $\mathbf{X_*}$ the joint distribution of the noise corrupted data $\mathbf{y}=f(\mathbf{X})+\varepsilon,\varepsilon \sim N(0,\sigma^2I)$ and $f(\mathbf{X_*})=\mathbf{f_*}$ is expressed as 

\begin{equation} \label{eq:app_GP_pred}
\begin{bmatrix}\mathbf{y} \\ \mathbf{f}_*\end{bmatrix} \sim \mathcal{N}
\left(\begin{bmatrix}\mathbf{\mu(\mathbf{X})} \\ \mathbf{\mu(\mathbf{X}_*)}\end{bmatrix},
\begin{bmatrix}\mathbf{K} +\sigma^2I & \mathbf{K}_* \\ \mathbf{K}_*^T & \mathbf{K}_{**}\end{bmatrix}
\right),
\end{equation}
where $\mathbf{K} = K(\mathbf{X,X}),$ $\mathbf{K_*} = K(\mathbf{X,X_*})$ and $\mathbf{K_{**}} = K(\mathbf{X_*,X_*}).$
The conditional distribution o $p(\mathbf{f}_* \mid \mathbf{X}_*,\mathbf{X},\mathbf{y})$ is also multivariate normal and more specifically

\begin{align*}
p(\mathbf{f}_* \mid \mathbf{X}_*,\mathbf{X},\mathbf{y}) &= \mathcal{N}(\boldsymbol{\mu}_*, \boldsymbol{\Sigma}_*) \\
\text{where } \boldsymbol{\mu_*} &= \boldsymbol{\mu(X_*)}+ \mathbf{K}_*^T (\mathbf{K}+\sigma^2 I)^{-1} (\mathbf{y}-\boldsymbol{\mu(X)})\\
\text{and } \boldsymbol{\Sigma_*} &= \mathbf{K}_{**} - \mathbf{K}_*^T (\mathbf{K}+\sigma^2 I)^{-1} \mathbf{K}_*\,.
\end{align*}.

\subsection{Physics-Informed priors prediction equations} \label{app:PE}
For the differential equation $\mathcal{L}_{x}^{\boldsymbol\phi}u(x) = f(x),$ by assuming that $u(x)\sim$ \\$GP(\mu_u(x), K_{uu}(x,x')),$  for the noisy corrupted data $\mathbf{y}_u = u(\mathbf{X}_u)+\boldsymbol{\varepsilon}_u, \boldsymbol{\varepsilon}_u \sim \mathcal{N}(0,\sigma_uI_u)$ and $\mathbf{y}_f = f(\mathbf{X}_f)+\boldsymbol{\varepsilon}_f, \boldsymbol{\varepsilon}_f \sim \mathcal{N}(0,\sigma_fI_f),$
we derive the physics-informed prior which is the following multi-output GP

\begin{equation}
  p(\mathbf{y}\mid \boldsymbol \theta,  \boldsymbol \phi, \sigma_u, \sigma_f) = \mathcal{N}(\boldsymbol{\mu}, \mathbf{K}+ \mathbf{S})
\end{equation}
where 
$
\bf{y} = \begin{bmatrix}  \bf{y}_u \\  \bf{y}_f \end{bmatrix},
$ 
$
\boldsymbol{\mu} = \begin{bmatrix}  \mu_u(\bf{X}_u) \\  \mu_f(\bf{X}_f) \end{bmatrix},
$
$
\mathbf{K} =
\begin{bmatrix}
K_{uu}(\mathbf{X}_u, \mathbf{X}_u \mid \boldsymbol \theta) & K_{uf}(\mathbf{X}_u, \mathbf{X}_f \mid \boldsymbol \theta,  \boldsymbol \phi)\\
K_{fu}(\mathbf{X}_f, \mathbf{X}_u \mid \boldsymbol \theta,  \boldsymbol \phi) & K_{ff}(\mathbf{X}_f, \mathbf{X}_f \mid \boldsymbol \theta,  \boldsymbol \phi) 
\end{bmatrix}
$ and
$
\mathbf{S} =
\begin{bmatrix}
\sigma_u^2 I_u & 0\\
0 &  \sigma_f^2 I_f
\end{bmatrix}. 
$\\
Applying the same logic as in eq. \ref{eq:app_GP_pred} at new points $\mathbf{X}_u^*$ we derive the prediction equations of $\mathbf{u}^* = u(\mathbf{X}_u^*)$ as follows

\begin{equation} 
\begin{bmatrix}\mathbf{y} \\ \mathbf{u}_*\end{bmatrix} \sim \mathcal{N}
\left(\begin{bmatrix}\mathbf{\mu(\mathbf{X})} \\ \mathbf{\mu(\mathbf{X}_u^*)}\end{bmatrix},
\begin{bmatrix}\mathbf{K} + \mathbf{S} & \mathbf{K}_* \\ \mathbf{K}_*^T & \mathbf{K}_{**}\end{bmatrix}
\right),
\end{equation}
For convenience we denote the vector of unknown parameters as $\boldsymbol{\xi} = (\boldsymbol \theta,  \boldsymbol \phi, \sigma_u, \sigma_f).$ The conditional distribution $p(\mathbf{u}_* \mid \mathbf{X}_u^*, \mathbf{X}, \mathbf{y}, \boldsymbol{\xi})$ is multivariate Gaussian and more specifically

\begin{align*}
p(\mathbf{u}_* \mid \mathbf{X}_u^*, \mathbf{X}, \mathbf{y}, \boldsymbol{\xi}) &= \mathcal{N}(\boldsymbol{\mu}_u^*, \boldsymbol{\Sigma}_u^*) \\
\boldsymbol{\mu_u^*} &= \mu_u\mathbf{(X_u^*)}+ \mathbf{V}_u^*{^T} (\mathbf{K}+\mathbf{S})^{-1} (\mathbf{y}-\boldsymbol{\mu})\\
\boldsymbol{\Sigma_u^*} &= K_{uu}(\mathbf{X}_u^*,\mathbf{X}_u^*) - \mathbf{V}_u^*{^T} (\mathbf{K}+\mathbf{S})^{-1} \mathbf{V}_u^*,
\end{align*}
where $\mathbf{V}_u^*{^T} = \begin{bmatrix}  K_{uu}(\mathbf{X}_u^*,\mathbf{X}_u) &  K_{uf}(\mathbf{X}_u^*,\mathbf{X}_f) \end{bmatrix}.$\\
Similarly, at new points $\mathbf{X}_f^*$ we derive the prediction equations of $\mathbf{f}^* = f(\mathbf{X}_f^*)$ as follows

\begin{equation} 
\begin{bmatrix}\mathbf{y} \\ \mathbf{f}_*\end{bmatrix} \sim \mathcal{N}
\left(\begin{bmatrix}\mathbf{\mu(\mathbf{X})} \\ \mathbf{\mu(\mathbf{X}_f^*)}\end{bmatrix},
\begin{bmatrix}\mathbf{K} & \mathbf{K}_* \\ \mathbf{K}_*^T & \mathbf{K}_{**}\end{bmatrix}
\right).
\end{equation}
The conditional distribution $p(\mathbf{f}_* \mid \mathbf{X}_f^*, \mathbf{X}, \mathbf{y}, \boldsymbol{\xi})$ is multivariate Gaussian and more specifically

\begin{align*}
p(\mathbf{f}_* \mid \mathbf{X}_f^*, \mathbf{X}, \mathbf{y}, \boldsymbol{\xi}) &= \mathcal{N}(\boldsymbol{\mu}_f^*, \boldsymbol{\Sigma}_f^*) \\
\boldsymbol{\mu_f^*} &= \mu_f\mathbf{(X_f^*)}+ \mathbf{V}_f^*{^T} (\mathbf{K}+\mathbf{S})^{-1} (\mathbf{y}-\boldsymbol{\mu})\\
\boldsymbol{\Sigma_f^*} &= K_{ff}(\mathbf{X}_f^*,\mathbf{X}_f^*) - \mathbf{V}_f^*{^T} (\mathbf{K}+\mathbf{S})^{-1} \mathbf{V}_f^*,
\end{align*}
where $\mathbf{V}_f^*{^T} = \begin{bmatrix}  K_{fu}(\mathbf{X}_f^*,\mathbf{X}_u) &  K_{ff}(\mathbf{X}_f^*,\mathbf{X}_f) \end{bmatrix}.$

% -------------------------------------------------------------------
\subsection{Accounting for model discrepancy prediction equations}\label{app:PE_disc}
By assuming a zero mean GP prior on the model discrepancy, $\delta(x)\sim GP(0, K_{\delta}(x,x'\mid \boldsymbol{\theta}_{\delta}))$ the model is similar to the Appendix \ref{app:PE} with the main difference that the discrepancy kernel is added to the first element of the covariance matrix $\mathbf{K}.$ More specifically, we have now that 

\begin{equation*}
    \mathbf{K}_{\text{disc}} =
    \begin{bmatrix}
    K_{uu}(\mathbf{X}_u, \mathbf{X}_u \mid \boldsymbol \theta) + K_{\delta}(\mathbf{X}_u, \mathbf{X}_u \mid \boldsymbol \theta_{\delta}) & K_{uf}(\mathbf{X}_u, \mathbf{X}_f \mid \boldsymbol \theta,  \boldsymbol \phi)\\
    K_{fu}(\mathbf{X}_f, \mathbf{X}_u \mid \boldsymbol \theta,  \boldsymbol \phi) & K_{ff}(\mathbf{X}_f, \mathbf{X}_f \mid \boldsymbol \theta,  \boldsymbol \phi) 
    \end{bmatrix}.
\end{equation*}
The vector of the parameters $\boldsymbol{\xi}$ has been augmented with the vector $\boldsymbol{\theta}_{\delta}$ and we denote all the kernel parameters collectively with $\boldsymbol{\xi}_{\delta} = (\boldsymbol \theta, \boldsymbol{\theta}_{\delta},  \boldsymbol \phi, \sigma_u, \sigma_f).$ Following the same logic as in the Appendix \ref{app:PE} we have that

\begin{align*}
p(\mathbf{u}_* \mid \mathbf{X}_u^*, \mathbf{X}, \mathbf{y}, \boldsymbol{\xi}_\delta) &= \mathcal{N}(\boldsymbol{\mu}_u^*, \boldsymbol{\Sigma}_u^*) \\
\boldsymbol{\mu_u^*} &= \mu_u\mathbf{(X_u^*)}+ \mathbf{V}_u^*{^T} (\mathbf{K_{\text{disc}}}+\mathbf{S})^{-1} (\mathbf{y}-\boldsymbol{\mu})\\
\boldsymbol{\Sigma_u^*} &= K_{uu}(\mathbf{X}_u^*,\mathbf{X}_u^*)+K_{\delta}(\mathbf{X}_u^*,\mathbf{X}_u^*) - \mathbf{V}_u^*{^T} (\mathbf{K_{\text{disc}}}+\mathbf{S})^{-1} \mathbf{V}_u^*,
\end{align*}
where $\mathbf{V}_u^*{^T} = \begin{bmatrix}  K_{uu}(\mathbf{X}_u^*,\mathbf{X}_u)+K_{\delta}(\mathbf{X}_u^*,\mathbf{X}_u) &  K_{uf}(\mathbf{X}_u^*,\mathbf{X}_f) \end{bmatrix}.$\\
The conditional distribution $p(\mathbf{f}_* \mid \mathbf{X}_f^*, \mathbf{X}, \mathbf{y}, \boldsymbol{\xi})$ is multivariate Gaussian and more specifically

\begin{align*}
p(\mathbf{f}_* \mid \mathbf{X}_f^*, \mathbf{X}, \mathbf{y}, \boldsymbol{\xi}_{\delta}) &= \mathcal{N}(\boldsymbol{\mu}_f^*, \boldsymbol{\Sigma}_f^*) \\
\boldsymbol{\mu_f^*} &= \mu_f\mathbf{(X_f^*)}+ \mathbf{V}_f^*{^T} (\mathbf{K_{\text{disc}}}+\mathbf{S})^{-1} (\mathbf{y}-\boldsymbol{\mu})\\
\boldsymbol{\Sigma_f^*} &= K_{ff}(\mathbf{X}_f^*,\mathbf{X}_f^*) - \mathbf{V}_f^*{^T} (\mathbf{K_{\text{disc}}}+\mathbf{S})^{-1} \mathbf{V}_f^*,
\end{align*}
where $\mathbf{V}_f^*{^T} = \begin{bmatrix}  K_{fu}(\mathbf{X}_f^*,\mathbf{X}_u) &  K_{ff}(\mathbf{X}_f^*,\mathbf{X}_f) \end{bmatrix}.$

% --------------------------------------------------------------------

\subsection{Accounting for Biased measurements prediction equations}\label{app:Biased_meas}

This case is similar to the model discrepancy case but here we want to remove the Bias in the model predictions. By assuming a zero mean GP prior on the Bias, $\text{Bias}(x)\sim GP(0, K_{\text{Bias}}(x,x'\mid \boldsymbol{\theta}_{B}))$ the model is similar to the Appendix \ref{app:PE_disc} with the difference that the discrepancy kernel, $\mathbf{K}_\delta$ is replaced by the Bias kernel $\mathbf{K}_{\text{Bias}}.$ More specifically, we have now that 

\begin{equation*}
    \mathbf{K}_{\text{Bias}} =
    \begin{bmatrix}
    K_{uu}(\mathbf{X}_u, \mathbf{X}_u \mid \boldsymbol \theta) + K_{\text{B}}(\mathbf{X}_u, \mathbf{X}_u \mid \boldsymbol \theta_B) & K_{uf}(\mathbf{X}_u, \mathbf{X}_f \mid \boldsymbol \theta,  \boldsymbol \phi)\\
    K_{fu}(\mathbf{X}_f, \mathbf{X}_u \mid \boldsymbol \theta,  \boldsymbol \phi) & K_{ff}(\mathbf{X}_f, \mathbf{X}_f \mid \boldsymbol \theta,  \boldsymbol \phi) 
    \end{bmatrix}. 
\end{equation*}
The vector of the parameters $\boldsymbol{\xi}$ has been augmented with the vector $\boldsymbol{\theta}_{\text{B}}$ and we denote all the kernel parameters collectively with $\boldsymbol{\xi}_{B} = (\boldsymbol \theta, \boldsymbol{\theta}_B,  \boldsymbol \phi, \sigma_u, \sigma_f).$ Following the same logic as in the Appendix \ref{app:PE} we have that

\begin{align*}
p(\mathbf{u}_* \mid \mathbf{X}_u^*, \mathbf{X}, \mathbf{y}, \boldsymbol{\xi}_B) &= \mathcal{N}(\boldsymbol{\mu}_u^*, \boldsymbol{\Sigma}_u^*) \\
\boldsymbol{\mu_u^*} &= \mu_u\mathbf{(X_u^*)}+ \mathbf{V}_u^*{^T} (\mathbf{K_{\text{Bias}}}+\mathbf{S})^{-1} (\mathbf{y}-\boldsymbol{\mu})\\
\boldsymbol{\Sigma_u^*} &= K_{uu}(\mathbf{X}_u^*,\mathbf{X}_u^*) - \mathbf{V}_u^*{^T} (\mathbf{K_{\text{Bias}}}+\mathbf{S})^{-1} \mathbf{V}_u^*,
\end{align*}
where $\mathbf{V}_u^*{^T} = \begin{bmatrix}  K_{uu}(\mathbf{X}_u^*,\mathbf{X}_u) &  K_{uf}(\mathbf{X}_u^*,\mathbf{X}_f) \end{bmatrix}.$\\
The conditional distribution $p(\mathbf{f}_* \mid \mathbf{X}_f^*, \mathbf{X}, \mathbf{y}, \boldsymbol{\xi})$ is multivariate Gaussian and more specifically

\begin{align*}
p(\mathbf{f}_* \mid \mathbf{X}_f^*, \mathbf{X}, \mathbf{y}, \boldsymbol{\xi}_B) &= \mathcal{N}(\boldsymbol{\mu}_f^*, \boldsymbol{\Sigma}_f^*) \\
\boldsymbol{\mu_f^*} &= \mu_f\mathbf{(X_f^*)}+ \mathbf{V}_f^*{^T} (\mathbf{K_{\text{Bias}}}+\mathbf{S})^{-1} (\mathbf{y}-\boldsymbol{\mu})\\
\boldsymbol{\Sigma_f^*} &= K_{ff}(\mathbf{X}_f^*,\mathbf{X}_f^*) - \mathbf{V}_f^*{^T} (\mathbf{K_{\text{Bias}}}+\mathbf{S})^{-1} \mathbf{V}_f^*,
\end{align*}
where $\mathbf{V}_f^*{^T} = \begin{bmatrix}  K_{fu}(\mathbf{X}_f^*,\mathbf{X}_u) &  K_{ff}(\mathbf{X}_f^*,\mathbf{X}_f) \end{bmatrix}.$

\section{Details on the physics-informed models}\label{app:PI_priors}
\subsection{Windkessel models}\label{app:WK}

\subsubsection*{\textbf{WK2 model}} 
The observed pressure, $y_P$ and inflow, $y_Q$ data are modelled by the physics-informed prior corrupted by Gaussian $i.i.d.$ noise $\varepsilon_P$ and $\varepsilon_Q$ respectively as follows

\begin{equation} \label{WK2_prior}
\begin{split}
	  y_P & = P^{\text{WK2}}(t_P) + \varepsilon_P\\
	  y_Q & = Q^{\text{WK2}}(t_Q) + \varepsilon_Q. 
\end{split}
\end{equation} 

To construct the physics-informed prior for the WK2 model we assume a GP prior on the pressure, $P^{\text{WK2}} \sim GP(\mu_P, K_{PP}(t,t')\mid \boldsymbol{\theta}).$ Then we have that 

\begin{equation} 
    \begin{split}
        K_{PQ}(t,t') &=  R^{-1} K_{PP}(t,t')+ C \frac{\partial K_{PP}(t,t')}{\partial t'}\\
        K_{QP}(t,t') &=  R^{-1} K_{PP}(t,t')+ C \frac{\partial K_{PP}(t,t')}{\partial t}\\
        K_{QQ}(t,t') &=  R^{-2} K_{PP}(t,t')+ C^2 \frac{\partial^2 K_{PP}(t,t')}{\partial t \partial t'}
    \end{split}
\end{equation}

This holds for the following three models where $K_{PP}$ is replaced by $K_\textrm{SE}, K_\textrm{RQ} \text{ and } K_\textrm{Per}.$ \\
\noindent \textbf{M1}. \emph{Squared Exponential Kernel (SE)}, $K_{\textrm{SE}}(t,t') = \sigma^2{ exp\left(- 0.5\,{ \left( 
{\frac {t-t'}{l}} \right) }^{2}\right)}$\\

\begin{equation} 
    \begin{split}
        R,C &\sim \mathcal{U}(0.5,3) \\
        \ell_{\text{WK2}} &\sim \text{Half-}\mathcal{N}(0, 1 / 3) \\
        \sigma_{\text{WK2}} &\sim \text{Half-}\mathcal{N}(0, 50) \\
        \sigma_P, \sigma_Q &\sim \text{Half-}\mathcal{N}(0, 15).
    \end{split}
\end{equation}

\noindent \textbf{M2}. \emph{Rational Quadratic Kernel (RQ)}, $K_{\textrm{RQ}}(t, t') = \sigma^2 \left( 1 + 
\frac{(t - t')^2}{2 \alpha \ell^2} \right)^{-\alpha}$\\
The same priors as the SE kernel are used with the addition of a uniform prior on $\alpha, $
$\alpha \sim \mathcal{U}(0, 10).$

\noindent \textbf{M3}. \emph{Periodic Kernel (Per)}, $K_{\textrm{Per}}(t,t') = \sigma^2\exp\left(-\frac{2\sin^2
(\pi(t - t')/p)}{\ell^2}\right)$\\
The same priors as the SE kernel are used for $R,C,\sigma_{\text{WK2}, \sigma_P \text{ and } \sigma_Q}$ with the addition of a uniform prior on $p,$ $p \sim \mathcal{U}(0.8, 1.2)$ and $\ell_{\text{WK2}}\sim \text{Half-}\mathcal{N}(0, 1)$

\subsubsection*{$\textbf{WK2}+\boldsymbol{\delta}(t)$ \textbf{model}}
The observed pressure now is described by the WK2 model and a functional model discrepancy, $\delta(t)$ corrupted by $i.i.d.$ noise as well, while the observed inflow, $y_Q$ is as before (eq. \ref{WK2_prior}) and more specifically

\begin{equation} 
\begin{split}
y_P & = P^{\text{WK2}}(t_P) + \delta(t_P) +\varepsilon_P\\
y_Q & = Q^{\text{WK2}}(t_Q) + \varepsilon_Q. 
\end{split}
\end{equation} 
The priors on $P^{\text{WK2}},$ the physical parameters $R,C$ and hyperparameters $\boldsymbol\theta$ are the same as in the WK2 models (M1,M2 and M3). In addition, we assume a GP prior on the model discrepancy, $\delta(t_P) \sim GP(0, K_{\delta}(t_P,t_P')\mid \boldsymbol{\theta_{\delta}}).$ The three following models are fitted:

\noindent For M1 (SE) and M2 (RQ) a squared exponential kernel is used for as kernel for the GP prior on the discrepancy function $\delta (t),$ where
\begin{equation} 
    \begin{split}
        \ell_{\delta} & \sim \text{Half-}\mathcal{N}(0, 1 / 3) \\
        \sigma_{\delta} & \sim \text{Half-}\mathcal{N}(0, 50).
    \end{split}
\end{equation}
\noindent For M3 (Per) a periodic kernel is used for as kernel for the GP prior on the discrepancy function $\delta (t),$ where
$\ell_{\delta} \sim \text{Half-}\mathcal{N}(0, 1),$ $\sigma_{\delta} \sim \text{Half-}\mathcal{N}(0, 50)$ and the same periodic parameter $p$ is used.

\subsection{Heat equation}\label{app:HeatEq}

\subsubsection*{$\mathbf{u(t,x)}$ \textbf{model}}
To develop the physics-informed prior we assume that the heat follows a GP prior, $u(t,x)\sim GP(\mu_u,K_{uu}((t,x), (t',x')))$ where we use an anisotropic squared exponential kernel, $K_{uu}((t,x), (t',x')) = \sigma^2 \exp\left(-\frac{1}{2l_t^2}(t-t')^2\right) \exp\left(-\frac{1}{2l_x^2}(x-x')^2\right)$ and $\mu$ is a constant. Then we have that 

\begin{equation} 
    \begin{split}
        K_{uf}((t,x), (t',x')) &=  \frac{\partial K_{uu}((t,x), (t',x'))}{\partial t'}- 
        \alpha \frac{\partial^2 K_{uu}((t,x), (t',x'))}{(\partial x')^2}\\
        K_{fu}((t,x), (t',x')) &=  \frac{\partial K_{uu}((t,x), (t',x'))}{\partial t}- 
        \alpha\frac{\partial^2 K_{uu}((t,x), (t',x'))}{(\partial x)^2}\\
        K_{ff}((t,x), (t',x')) &=  \frac{\partial^2 K_{uu}((t,x), (t',x'))}{\partial t\partial t'}+ 
        \alpha^2 \frac{\partial^4 K_{uu}((t,x), (t',x'))}{(\partial x )^2(\partial x')^2}
    \end{split}
\end{equation}
We use the following weakly informative priors:
\begin{equation} 
    \begin{split}
        \alpha &\sim \mathcal{U}(0,10) \\
        \ell_x &\sim \text{Half-}\mathcal{N}(0, 1 / 3) \\
        \ell_t &\sim \text{Half-}\mathcal{N}(0, 1 ) \\
        \sigma &\sim \text{Half-}\mathcal{N}(0, 1 / 3) \\
        \mu & \sim \text{Half-}\mathcal{N}(0.5, 1 ) \\
        \sigma_u & \sim \mathcal{U}(0,0.5) \\
        \sigma_f & \sim \mathcal{U}(0,3).\\
    \end{split}
\end{equation}

\subsubsection*{$\mathbf{u(t,x)} + \textbf{Bias}\mathbf{(t,x)}$ \textbf{model}}

The model priors are the same as for the $u(t,x)$ model. In addition, we assume a GP prior on Bias, $\mathrm{Bias}(t,x) \sim GP(0, K_{\mathrm{Bias}}((t,x),(t',x')))$ with an anisotrpic squared exponential kernel,  $K_{\mathrm{Bias}}((t,x),(t',x'))) = \sigma_B^2 exp\left(-\frac{1}{2lB_t^2}(t-t')^2\right) exp\left(-\frac{1}{2lB_x^2}(x-x')^2\right).$ The Bias kernel hyper-parameter priors are
\begin{equation} 
    \begin{split}
        \ell B_x &\sim \text{Half-}\mathcal{N}(0, 1 / 3) \\
        \ell B_t &\sim \text{Half-}\mathcal{N}(0, 1 ) \\
        \sigma B &\sim \text{Half-}\mathcal{N}(0, 1 / 3).\\
    \end{split}
\end{equation}

% --------------------------------
\bibliography{main}
\end{document}